%% file: main.tex
\documentclass[journal]{IEEEtran}
\IEEEoverridecommandlockouts
\usepackage{cite}
\usepackage{amsmath,amssymb,amsfonts}
\usepackage[ruled,vlined]{algorithm2e}
\usepackage{graphicx}
\usepackage{textcomp}
\usepackage{xcolor}
\usepackage{subcaption}
\usepackage{multicol}
\usepackage{multirow}
\usepackage{url}
\usepackage{hyperref}
\def\BibTeX{{\rm B\kern-.05em{\sc i\kern-.025em b}\kern-.08em
    T\kern-.1667em\lower.7ex\hbox{E}\kern-.125emX}}

\definecolor{darkgreen}{RGB}{0,0,0}
\newcommand{\green}[1]{{\color{darkgreen}#1}}
\definecolor{darkgreen2}{RGB}{0,120,0}

\begin{document}
\bstctlcite{IEEEexample:BSTcontrol}

%%%REMOVE BEFORE SUBMITTING
%\thispagestyle{plain}
%\pagestyle{plain}

%\title{A Spatio-Temporal Scene-Graph Embedding Approach for AV Collision Prediction\\
\title{Spatio-Temporal Scene-Graph Embedding for Autonomous Vehicle Collision Prediction
}

%\author{Anonymous Authors}
\author{
    \IEEEauthorblockN{Arnav V. Malawade*\thanks{* Arnav V. Malawade and Shih-Yuan Yu contributed equally to this article.}, Shih-Yuan Yu*, Brandon Hsu, Deepan Muthirayan, \\Pramod~P.~Khargonekar, 
    Mohammad A. Al Faruque}\\
    % \IEEEauthorblockN{Shih-Yuan Yu\IEEEauthorrefmark{1}, Arnav V. Malawade\IEEEauthorrefmark{1}, Deepan Muthirayan\IEEEauthorrefmark{1}, Pramod~P.~Khargonekar\IEEEauthorrefmark{1}, 
    % Mohammad A. Al Faruque\IEEEauthorrefmark{1}}\\
    \IEEEauthorblockA{Department of Electrical Engineering \& Computer Science, University of California - Irvine, Irvine, CA 92697, USA}
    \IEEEauthorblockA{\{malawada, shihyuay, bdhsu, dmuthira, pramod.khargonekar, alfaruqu\}@uci.edu }
 }%

\maketitle

\input{tex/abstract}

\begin{IEEEkeywords}
scene graph, collision prediction, autonomous vehicles, graph learning, ADAS
\end{IEEEkeywords}

\input{tex/intro-v1}
\input{tex/related_work}
\input{tex/methodology}
\input{tex/results}
\input{tex/conclusion}

\bibliographystyle{IEEEtran}
\bibliography{IEEEabrv, references}

%add bios here
\input{tex/author_bios}

\end{document}

%% file: tex/abstract.tex
\begin{abstract}
In autonomous vehicles (AV), early warning systems rely on collision prediction to ensure occupant safety.
However, state-of-the-art methods using deep convolutional networks either fail at modeling collisions or are too expensive/slow, making them less suitable for deployment on AV edge hardware.
To address these limitations, we propose \textsc{sg2vec}, a spatio-temporal \textit{scene-graph} embedding methodology that uses Graph Neural Network (GNN) and Long Short-Term Memory (LSTM) layers to predict future collisions via visual scene perception.
We demonstrate that \textsc{sg2vec} predicts collisions 8.11\% more accurately and 39.07\% earlier than the state-of-the-art method on synthesized datasets, and 29.47\% more accurately on a challenging real-world collision dataset.
We also show that \textsc{sg2vec} is better than the state-of-the-art at transferring knowledge from synthetic datasets to real-world driving datasets.
Finally, we demonstrate that \textsc{sg2vec} performs inference 9.3x faster with an 88.0\% smaller model, 32.4\% less power, and 92.8\% less energy than the state-of-the-art method on the industry-standard Nvidia DRIVE PX 2 platform, making it more suitable for implementation on the edge.
\end{abstract}

%100 word version:
% Autonomous vehicles rely on collision prediction to ensure occupant safety. However, state-of-the-art methods either fail at modeling collisions or are too expensive/slow, making them unsuitable for edge implementation. We propose sg2vec: a scene-graph embedding methodology that uses graph neural networks to predict future collisions. We demonstrate that sg2vec predicts collisions better and earlier than the state-of-the-art method. We also show that sg2vec better transfers knowledge from synthetic to real-world datasets. Finally, we demonstrate that sg2vec performs faster inference while using less power/energy on the industry-standard Nvidia DRIVE PX 2 platform, making it more suitable for implementation on the edge.

%% file: tex/intro-v1.tex
\section{Introduction}
\label{sec:intro}
The synergy of Artificial Intelligence (AI) and the Internet of Things (IoT) has accelerated the advancement of Autonomous Vehicle (AV) technologies, which is expected to revolutionize transportation by reducing traffic and improving road safety~\cite{vom2020fail, bijlsma2020distributed}. 
%motivation:
However, recent reports of AV crashes suggest that there are still significant limitations. For example, multiple fatal Tesla Autopilot crashes can primarily be attributed to perception system failures \cite{NTSB2018, NTSB2019}. Additionally, the infamous fatal collision between an Uber self-driving vehicle and a pedestrian can be attributed to perception and prediction failures by the AV \cite{NTSB2019uber}. 
These accidents (among others) have eroded public trust in AVs, and nearly 50\% or more of the public have expressed their mistrust in AVs \cite{deloitte2020}. 
Current statistics indicate that perception and prediction errors were factors in over 40\% of driver-related crashes between conventional vehicles \cite{mueller2020humanlike}. 
However, a significant number of reported AV collisions are also the result of these errors \cite{schoettle2015preliminary, xu2019statistical}.
%Thus, in this paper, we aim to improve the safety and acceptance of AVs by targeting the specific task of early collision prediction.
Thus, in this paper, we aim to improve the safety and acceptance of AVs by incorporating \textit{scene-graphs} into the perception pipeline of collision prediction systems to improve scene understanding.

%industrial SOTA collision avoidance and limitations:
For the past several years, automotive manufacturers have begun equipping consumer vehicles with statistics-based collision avoidance systems based on calculated Single Behavior Threat Metrics (SBTMs) such as Time to Collision (TTC), Time to React (TTR), etc. \cite{volvo2012safety, dahl2018collision}. 
However, these methods lack robustness since they make significant assumptions about the behavior of vehicles on the road. A very limiting assumption they make is that vehicles do not diverge from their current trajectories~\cite{dahl2018collision}.
SBTMs can also fail in specific scenarios. For example, TTC can fail when following a vehicle at the same velocity within a very short distance~\cite{dahl2018collision}.
As a result, these methods are less capable of generalizing and can perform poorly in complex road scenarios.
Moreover, to reduce false positives, these systems are designed to respond at the last possible moment \cite{sontges2018worst}. Under such circumstances, the AV control system can fail to take timely corrective actions \cite{nilsson2015worst} if the system fails to predict a collision or estimates the TTC inaccurately.

%DL approaches for collision avoidance:
More effective collision prediction methods using Deep Learning (DL) have also been proposed in the literature. 
However, these approaches can be limited because they do not explicitly capture the relationships between the objects in a traffic scene. Understanding these relationships could be critical as it is suggested that a human's ability to understand complex scenarios and identify potential risks relies on cognitive mechanisms for representing structure and reasoning about inter-object relationships~\cite{battaglia2018relational}. These models also require large datasets that are often costly or unsafe to generate. Synthetic datasets are typically used to augment the limited real-world data to train the models in such cases~\cite{dosovitskiy2017carla}.
However, these trained models must then be able to transfer the knowledge gained from synthetic datasets to real-world driving scenarios.
Furthermore, DL models contain millions of parameters and require IoT edge devices with significant computational power and memory to run efficiently. 
Likewise, hosting these models on the cloud is infeasible because it requires persistent low-latency internet connections.
% Thus, an ideal collision prediction model should operate effectively within the constraints of the AV CPU without compromising the performance of other critical functions (control, perception, planning, etc.)
% However, these methods have limitations in road scenarios with complex traffic, environmental, and road conditions. 

In summary, the key research challenges associated with autonomous vehicle collision prediction are:
\begin{enumerate}
    \item Capturing complex relationships between road participants.
    \item Detecting future collisions early enough such that the AV can take corrective actions.
    \item Generalizing to a wide range of traffic scenarios.
    \item Developing algorithms that can run efficiently on automotive IoT edge devices. 
\end{enumerate} 
%\begin{enumerate}
%    \item {\it Capturing the complex relationships between road participants}.
%    \item {\it Detecting future collisions early enough that the AV can take corrective actions}.
%    \item {\it Generalizing to a wide range of traffic scenarios}.
%    \item {\it Developing algorithms which can run effectively on the edge}.
%\end{enumerate}

\begin{figure}[t]
    \centering
    \includegraphics[clip, trim=95 25 315 30, width=\linewidth]{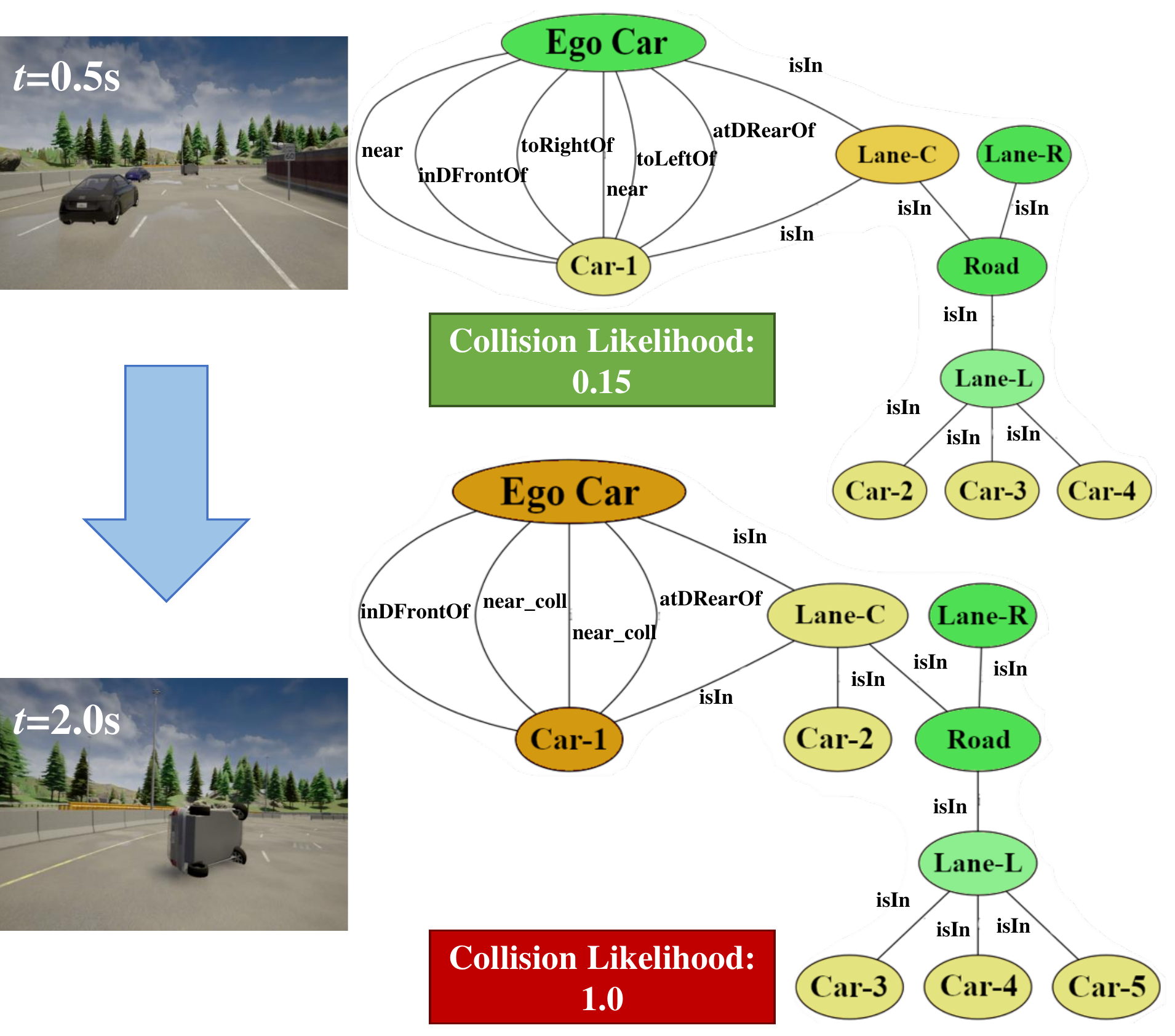}
    \caption{How \textsc{sg2vec} predicts collisions using \textit{scene-graphs}. Each node's color indicates its attention score (importance to the collision likelihood) from orange (high) to green (low).}
    \label{fig:motivation}
    %\vspace{-1.5em}
\end{figure}

In this work, we propose the use of \textit{scene-graphs} to represent road scenes and model inter-object relationships as we hypothesize that this will improve perception and scene understanding. Recently, several works have shown that graph-based methods that capture and model complex relationships between entities can improve performance at high-level tasks such as behavior classification \cite{mylavarapu2020towards, li2020learning} and semantic segmentation \cite{kunze2018reading}. 
\textit{Scene-graphs} are used in many domains to abstract the locations and attributes of objects in a scene and model inter-object relationships~\cite{li2020learning, mylavarapu2020towards, mylavarapu2020understanding}. % \textit{Scene-graphs} are used in many domains to encode an entire scene modality into a graphical representation where the graph models the location and attributes of objects within the scene along with their pair-wise relations. <- If using this version, change below form of abstraction to form of representation
In our prior work, \cite{yu2021scene}, we demonstrated that a \textit{scene-graph} sequence classification approach can better assess the subjective risk of driving clips compared to a conventional CNN+LSTM-based approach; our approach could also better generalize and transfer knowledge across datasets. This paper extends the approach presented in \cite{yu2021scene} to collision prediction by enabling the prediction of future road states via changes to the temporal modeling components of the architecture and changes in the problem formulation.
The \textit{scene-graph} representation we propose represents traffic objects as nodes and the relationships between them as edges. The novelty of our \textit{scene-graph} representation lies in our graph construction technique that is specifically designed for higher-level AV decisions such as collision prediction. One of our key contributions is showing that {\it a graph-based intermediate representation can be very effective and efficient for higher-level AV tasks directly related to driving decisions}.

Our proposed methodology for collision prediction, \textsc{sg2vec}, is shown in Figure \ref{fig:motivation}. It combines the \textit{scene-graph} representation with a graph-embedding architecture to generate a sequence of \textit{scene-graph} embeddings for the sequence of visual inputs perceived by an AV. The graph embedding technique we use is based on the core MR-GCN framework \cite{mylavarapu2020towards} adapted for the collision prediction problem. The sequence of graph embeddings is then input to a Long Short-Term Memory (LSTM) network to make the final prediction on the possibility of a future collision. To the best of our knowledge, our work is the first to propose using \textit{scene-graphs} for early collision prediction. 

%key contributions
Our paper makes the following key research contributions:
\begin{enumerate}
%    \item {\it We demonstrate the performance of our \textsc{sg2vec} collision prediction methodology on simulated lane-change datasets and a very challenging real-world collision dataset containing a wide range of driving actions, collision types, and weather/road conditions}.
%    \item {\it We compare \textsc{sg2vec} with a state-of-the-art data-driven methodology on the metrics: accuracy, inference speed, and average time of prediction, among others}.
    \item {We demonstrate that our \textsc{sg2vec} collision prediction methodology significantly outperforms the current state of the art on simulated lane-change datasets and a very challenging real-world collision dataset containing a wide range of driving actions, collision types, and weather/road conditions.}
    \item {We demonstrate that \textsc{sg2vec} can transfer knowledge gained from simulated data to real-world driving data more effectively than the state-of-the-art method}.
    \item {We show that \textsc{sg2vec} performs faster inference and requires less power than the state-of-the-art method on the industry-standard Nvidia DRIVE PX 2 autonomous driving hardware platform, used in all 2016-2018 Tesla models for their Autopilot system \cite{px2tesla}}.
\end{enumerate} 

%% file: tex/related_work.tex
\section{Related Work}
%collision prediction references.
\subsection{Early Collision Prediction}
Since collision prediction is key to the safety of AVs, a wide range of solutions have been proposed by academia and industry.
As mentioned earlier, current consumer vehicles use statistics-based SBTMs for collision prediction but can perform poorly in complex situations \cite{volvo2012safety,dahl2018collision} or react too late to avoid collisions \cite{nilsson2015worst, sontges2018worst}. Expanding on these approaches, companies like Mobileye and Nvidia have proposed more comprehensive mathematical models for ensuring AV safety, namely Responsibility-Sensitive Safety (RSS) \cite{shalev2017formal} and Nvidia Safety Force Field \cite{nister2019safety}, respectively. However, these models are heavily rule-based and can thus be fragile in complex situations with high uncertainty. Additionally, computing future trajectory constraints %(i.e., avoiding collisions using both lateral (steering) and longitudinal corrections(braking)) 
with RSS is non-trivial and can require vehicle-specific calibration  \cite{gassmann2020integration}. %real-time feasibility and practicality \cite{gassmann2020integration}}. %This also means it is difficult for RSS to avoid collisions in complex scenarios that require more evasive maneuvers than just braking.}

Model-based probabilistic and deep learning approaches for collision prediction have also been proposed. For example, \cite{althoff2009model} proposes a model-based probabilistic technique that uses the roadway geometry, ego trajectory, and position/velocity of road objects to predict future object positions. However, this model is highly conservative and is likely to have a high false-positive rate. 
Similarly, \cite{wang2019trajectory} and \cite{zhang2020surrounding} use model-based approaches but require significant domain knowledge about the driving scene, such as road geometry information as well as accurate vehicle position and velocity information. 
\cite{wang2020real} proposes a deep learning collision prediction approach. Still, due to its use of pre-processed trajectory data captured from cameras overlooking a highway, it is not ego-centric and cannot be practically used for on-vehicle collision prediction.
In a different approach, \cite{strickland2018deep} proposes a Deep Predictive Model (DPM) that used a Bayesian Convolutional LSTM for collision risk assessment where image data, vehicle telemetry data, and driving inputs were all factors in the risk assessment decision. 
However, this approach was only evaluated on simulated street scenes containing two vehicles and no other dynamic objects. Thus, DPM's performance may suffer when evaluated on more complex road scenarios. 
%, such as those with variable weather/lighting conditions, a large number of vehicles, and multiple types of driving maneuvers and collisions. 

In contrast to these existing works, we propose \textsc{sg2vec} which captures structural and relational information of a road scene in a \textit{scene-graph} representation and computes a spatio-temporal embedding to predict collisions. 
Additionally, we perform experiments that were not done in many prior works, such as evaluating each model's capability to transfer knowledge, efficiency on AV hardware, performance on a complex real-world crash dataset, and ability to predict collisions early.
We primarily compare our methodology with the DPM as it is the state-of-the-art data-driven collision prediction framework for AVs that considers both spatial and temporal factors. 
Although the DPM uses multiple modalities for sensing, the results in \cite{strickland2018deep} show that it achieves an accuracy (of 81.95\%) that is just 0.24\% less using just the image sensing modality. In this work, we compare our proposed \textsc{sg2vec} methodology and the DPM on image-only datasets, which is fair because the DPM's performance does not vary much with the inclusion of other modalities.

%scene-graph related works.
\subsection{AV Scene-Graphs and Optimization Techniques}
Several works have proposed graph-based methods for scene understanding. For example, \cite{mylavarapu2020towards} proposed a multi-relational graph convolutional network (MR-GCN) that uses both spatial and temporal information to classify vehicle driving behavior. 
%The MR-GCN proposed in this work observes how the directional relations between moving objects and fixed landmarks change over time. 
Similarly, in \cite{li2020learning}, an \textit{Ego-Thing} and \textit{Ego-Stuff} graph are used to model and classify the ego vehicle's interactions with moving and stationary objects, respectively.
In our prior work, we demonstrated that a \textit{scene-graph} sequence embedding approach assesses driving risk better than the state-of-the-art CNN-LSTM approach~\cite{yu2021scene}. In \cite{yu2021scene}, we utilized an architecture consisting of MR-GCN layers for spatial modeling and an LSTM with attention for temporal modeling; however, this architecture was only capable of performing binary sequence-level classification over a complete video clip. Thus, although our prior architecture could accurately assess the subjective risk of complete driving sequences, it was not capable of predicting the future state of a scene. 
%Besides, we also found that our approach can provide explainable predictions and can effectively transfer knowledge learned from synthetic datasets to real-driving scenarios~\cite{yu2021scene}.

%DAC/DATE references.
%\subsection{Autonomous Vehicle Power/Energy Optimization}
Current autonomous driving systems consume a substantial amount of power (up to 500 Watts for the Nvidia DRIVE AGX Pegasus), demanding more robust cooling and power delivery mechanisms. 
Thus, many have tried to optimize AV tasks for efficiency without sacrificing performance. Existing approaches have proposed methods for jointly optimizing power consumption and latency for localization \cite{asgari2020pisces}, perception \cite{baidya2020vehicular}, and control \cite{huang2020opportunistic}.  However, to the best of our knowledge, no work has explored this optimization for AV safety systems, such as collision prediction systems. 

%% file: tex/methodology.tex
\begin{figure}[t]
\centering
    \includegraphics[clip, trim=120 125 240 55, width=\linewidth]{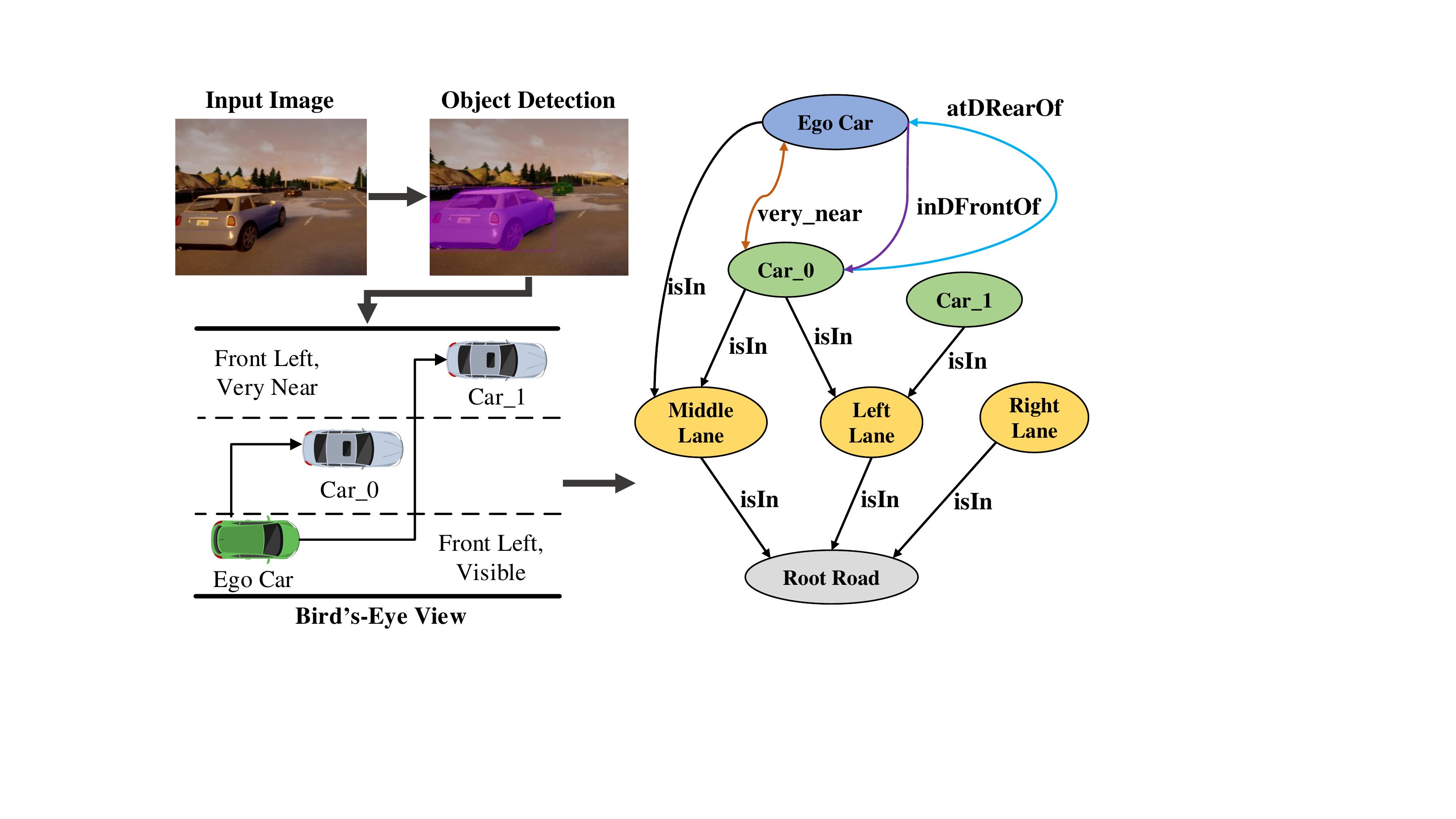}
    \caption{An illustration of our \textit{scene-graph} extraction process.}
    \label{fig:graphextraction}
    %\vspace{-1.5em}
\end{figure}

\begin{figure*}[t]
\centering
    \includegraphics[clip, trim=12 155 12 192, width=\textwidth]{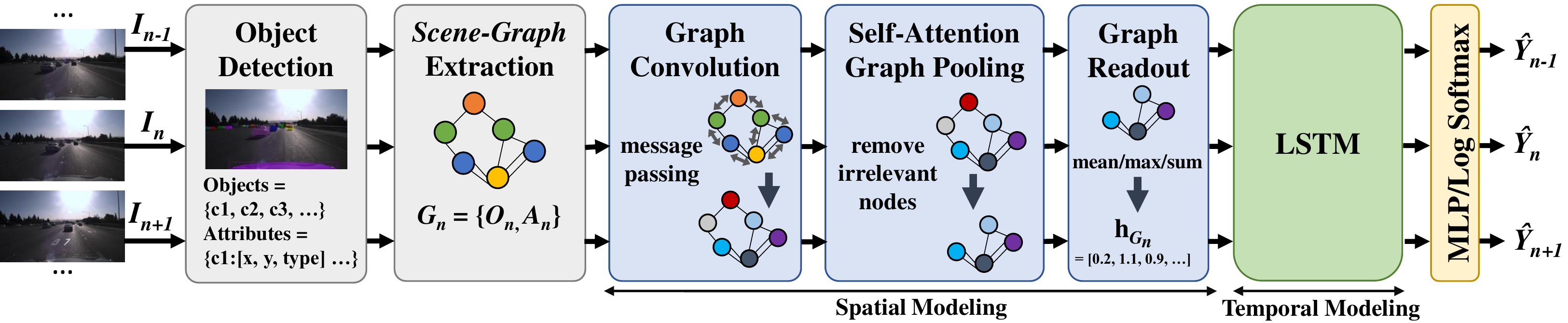}
    \caption{An illustration of \textsc{sg2vec}'s architecture.}
    \label{fig:archi}
    %\vspace{-1.5em}
\end{figure*}

\section{Scene-Graph Embedding Methodology}
In \textsc{sg2vec}, we formulate the problem of collision prediction as a time-series classification problem where the goal is to predict if a collision will occur in the \textit{near future}. 
Our goal is to accurately model the spatio-temporal function $f$, where
\begin{equation}
    \mathbf{Y_n} = f(\{I_{1}, ..., I_{n-1}, I_n\}), \mathbf{Y_n} \in \{0,1\}, \text{for } n > 2,
\end{equation}
%$\mathbf{Y_n}$ is the model's prediction of the likelihood of a collision in the near future, 
where $\mathbf{Y_n}=1$ implies a collision in the near future and $\mathbf{Y_n} = 0$ otherwise.
Here the variable $I_n$ denotes the image captured by the on-board camera at time $n$. The interval between each frame varies with the camera sampling rate. %Next, we describe our \textit{scene-graph} extraction process.
%In the context of this application, \textit{near future} refers to a time  of $1$ to $2$ seconds ahead of the current time $n$.

\textsc{sg2vec} consists of two parts (Figure~\ref{fig:archi}) : (i) the \textit{scene-graph} extraction, and (ii) collision prediction through spatio-temporal embedding, described in Section~\ref{sec:extraction} and Section~\ref{subsec:arch} respectively.
\subsection{Scene-Graph Extraction}
\label{sec:extraction}
%Existing works propose to extract \textit{scene-graphs} from images by detecting objects in a scene and then reasoning about their visual relationships ~\cite{yang2018graph,xu2017scene}. However, these works focus on leveraging the object-relationship regularities to infer the semantic relationships between objects, while \textsc{sg2vec} aims to use these relations to model the state of a traffic scene for higher-level tasks such as risk assessment and collision avoidance. Additionally, \textsc{sg2vec} only requires a subset of relations that are domain-specific to autonomous driving, such as directional relations and proximity relations. Therefore, \textsc{sg2vec} adopts a partially rule-based graph extraction pipeline to construct the \textit{scene-graph} as shown in Figure~\ref{fig:graphextraction}.

The first step of our methodology is the extraction of \textit{scene-graphs} for the images of a driving scene. The extraction pipeline forms the \textit{scene-graph} for an image as in ~\cite{yang2018graph,xu2017scene} by first detecting the objects in the image and then identifying their relations based on their attributes. The difference from prior works lies in the construction of a \textit{scene-graph} that is designed for higher-level AV decisions. We propose extracting a {\it minimal} set of relations such as directional relations and proximity relations. 
From our design space exploration we found that adding many relation edges to the \textit{scene-graph} adds noise and impacts convergence while using too few relation types reduces our model's expressivity. The best approach we found across applications involves constructing mostly ego-centric relations for a moderate range of relation types.
%We show later that the minimal \textit{scene-graph} construction we propose for this problem is very effective across datasets.
Figure~\ref{fig:graphextraction} shows an example of the graph extraction process.  

We denote the extracted \textit{scene-graph} for the frame $I_n$ by $G_n =\{O_n, A_n\}$.
Each \textit{scene-graph} $G_n$ is a directed, heterogeneous multi-graph, where $O_n$ denotes the nodes and $A_n$ is the adjacency matrix of the graph $G_n$. As shown in Fig.~\ref{fig:graphextraction}, nodes represent the identified objects such as lanes, roads, traffic signs, vehicles, pedestrians, etc., in a traffic scene. The adjacency matrix $A_n$ indicates the pair-wise relations between each object in $O_n$. The extraction pipeline first identifies the objects $O_n$ by using Mask R-CNN~\cite{he2017mask}. Then, it generates an inverse perspective mapping (also known as a ``birds-eye view" projection) of the image to estimate the locations of objects relative to the ego car, which are used to construct the pair-wise relations between objects in $A_n$. For each camera angle, we calibrate the birds-eye view projection settings using known fixed distances, such as the lane length and width, as defined by the highway code. This enables us to estimate longitudinal and lateral distances accurately in the projection. For datasets captured by a single vehicle, this step only needs to be performed once. However, for datasets with a wide range of camera angles such as the \textit{620-dash} dataset introduced later in the paper, this process needs to be performed once per vehicle. With a human operator, we found that this calibration step takes approximately 1 minute per camera angle on average.
%The matrix $A_n$ is then constructed by identifying the pair-wise relations between objects using their attributes (e.g., relative distance/position to one another).

% Originally before directional relation
% Here the assumption is that one object influences the other object's motion only if they are within a certain distance ($4,7,10,16,25$ ft. respectively).

%Here the assumption is that only the local proximity and positional information of one object could influence the other object's motion only if they are within a certain distance.
The extraction pipeline identifies three kinds of pair-wise relations: \textit{proximity} relations (e.g. \textit{visible}, \textit{near}, \textit{very\_near}, etc.), \textit{directional} (e.g. \textit{Front\_Left}, \textit{Rear\_Right}, etc.) relations, and \textit{belonging} (e.g. car\_1 \textit{isIn} left\_lane) relations. 
Two objects are assigned the \textit{proximity} relation, $r\in$ \{\textit{Near\_Collision} (4 ft.), \textit{Super\_Near} (7 ft.), \textit{Very\_Near} (10 ft.), \textit{Near} (16 ft.), \textit{Visible} (25 ft.)\} provided the objects are physically separated by a distance that is within that relation's threshold. 
The {\it directional relation}, $r \in$ \{\textit{Front\_Left}, \textit{Left\_Front}, \textit{Left\_Rear}, \textit{Rear\_Left}, \textit{Rear\_Right}, \textit{Right\_Rear}, \textit{Right\_Front}, \textit{Front\_Right}\}, is assigned to a pair of objects, in this case between the ego-car and another car in the view, based on their relative orientation and only if 
% Here, the directional relation is assigned only if 
they are within the \textit{Near} threshold distance from one another. 
Additionally, the \textit{isIn} relation identifies which vehicles are on which lanes (see Fig.~\ref{fig:graphextraction}). We use each vehicle's horizontal displacement relative to the ego vehicle to assign vehicles to either the \textit{Left Lane}, \textit{Middle Lane}, or \textit{Right Lane} using the known lane width. 
Our abstraction only considers three-lane areas, and, as such, we map vehicles in all left lanes and all right lanes to the same \textit{Left Lane} node \textit{Right Lane} node respectively. 
If a vehicle overlaps two lanes (i.e., during a lane change), it is mapped to both lanes.
\vspace{-1.0em}
% Our abstraction only considers three-lane areas, and, as such, we map vehicles in all left lanes to the same \textit{Left Lane} node and all vehicles in right lanes to the \textit{Right Lane} node. 
% If a vehicle overlaps two lanes (i.e., during a lane change), it is mapped to both lanes.
% \vspace{-1.0em}

\subsection{Collision Prediction}
\label{subsec:arch}
As shown in Figure \ref{fig:archi}, in our collision prediction methodology, each image $I_n$ is first converted into a \textit{scene-graph} $G_n = \{O_n, A_n\}$ with the pipeline mentioned in Section~\ref{sec:extraction}. 
Each node $v \in O_n$ is initialized by a one-hot vector (\textit{embedding}), denoted by $\mathbf{h}^{(0)}_v$.
Then, the MR-GCN~\cite{schlichtkrull2018modeling} layers are used to update these embeddings via the edges in $A_n$.
Specifically, the $l$-th MR-GCN layer computes the node embedding for each node $v$, denoted as $\mathbf{h}^{(l)}_v$, as follows:
\begin{equation}
    \mathbf{h}^{(l)}_v = \mathbf{\Phi}_{\textrm{0}} \cdot
        \mathbf{h}^{(l-1)}_v + \sum_{r \in \mathbf{A_{n}}} \sum_{u \in \mathbf{N}_r(v)}\frac{1}{|\mathbf{N}_r(v)|} \mathbf{\Phi }_r \cdot \mathbf{h}^{(l-1)}_u,
\end{equation}
where $N_r(v)$ denotes the set of neighbors of node $v$ with respect to the relation $r\in A_{n}$, $\mathbf{\Phi }_r$ is a trainable relation-specific transformation for relation $r$, and $\mathbf{\Phi }_0$ is the self-connection for each node $v$ that accounts for the influence of $\mathbf{h}^{(l-1)}_v$ on $\mathbf{h}^{(l)}_v$~\cite{schlichtkrull2018modeling}. After multiple MR-GCN layers, the output of each layer is concatenated to produce the final embedding for each node, denoted by $\mathbf{H}^{L}_v = \textbf{CONCAT}(\{\mathbf{h}^{(l)}_v\}|l=0, 1, ...,L)$, where $L$ is the index of the last layer. % \red{give more details on $H_v^L$ here. R2-1}

The final embeddings for \textit{scene-graph} $G_n$, denoted by $\mathbf{X}^{prop}_{n}$, are then passed through a self-attention graph pooling (SAGPooling) layer that filters out irrelevant nodes from the graph, creating the pooled set of node embeddings $\mathbf{X}^{pool}_{n}$ and their edges $\mathbf{A}^{pool}_{n}$. 
In this layer, we use a graph convolution layer to predict the score $\alpha = \mathbf{SCORE}(\mathbf{X}^{prop}_{n}, \mathbf{A}^{prop}_{n})$ and then use $\alpha$ to perform \textit{top-k} filtering to filter out the irrelevant nodes in the \textit{scene-graph} \cite{lee2019self}. % \red{give more details on $\alpha$ here. R2-1}

Then, for each \textit{scene-graph} $G_n$, the corresponding $\mathbf{X}^{pool}_{n}$ is passed through the graph readout layer that condenses the node embeddings (using operations such as sum, mean, max, etc.) to a single graph embedding $\mathbf{h}_{G_n}$. % \red{give more details on $h_{G_n}$ here and how its passed to LSTM to generate $z_n$. explain contribution of LSTM. R2-2}
%We denote the sequence of \textit{scene-graph} embeddings generated from each frame $I_n$ in a video clip of length $T$ as $h_{\mathbf{I}} = \{\mathbf{h}_{G_n}|n=1,..T\}$.
%The sequence of embeddings $h_{\mathbf{I}}$ is then passed to the temporal model that uses an LSTM layer to process the embeddings in $h_{\mathbf{I}}$ as a sequence. 
Then, this spatial embedding $\mathbf{h}_{G_n}$ is passed to the temporal model (LSTM) to generate a spatio-temporal embedding $z_n$ as follows:
\begin{equation}
    z_n, s_n = \mathbf{LSTM}(h_{G_n}, s_{n-1})
\end{equation}
The hidden state $s_n$ of the LSTM is updated for each timestamp $n$.
%The sequence of outputs of the temporal model $\mathbf{z} = \{z_{n}|n=1,2,...T\}$ is then be passed to an MLP layer to generate the inference $\hat{Y_n}$ for each timestamp $n$
Lastly, each spatio-temporal embedding $z_{n}$ is then passed through a Multi-Layer Perceptron (MLP) that outputs each class's confidence value.
The two outputs of the MLP are compared, and $\hat{Y}_n$ is set to the index of the class with the greater confidence value (0 for no-collision or 1 for collision).
During training, we calculate the cross-entropy loss between each set of non-binarized outputs $\hat{Y}_n$ and the corresponding labels for backpropagation.

% \begin{algorithm}
% \SetAlgoLined
% \LinesNumbered
% \DontPrintSemicolon
% \KwResult{Future Collision Likelihood: $\hat{Y}_n$}
% initialize $n \gets$ 1\ ,$s_0 \gets $ [0, 0, ..., 0]; \\
% \SetKwProg{Fn}{def}{:}{}
% \SetKwFunction{Fsgvec}{sg2vec}
% \SetKwFunction{FMain}{main}
% \Fn{\Fsgvec{$I_n$, $s_{n-1}$}}{
% $G_n \gets$ SceneGraphExtraction($I_n$)\;
% %$h_{G_n} \gets$ MR-GCN($G_n$)\;
% $\mathbf{X}^{prop}_{n}, \mathbf{A}^{prop}_{n} \gets$ MR-GCN($G_n$)\;
% $\mathbf{X}^{pool}_{n}, \mathbf{A}^{pool}_{n} \gets$ SAGPooling($\mathbf{X}^{prop}_{n}$, $ \mathbf{A}^{prop}_{n}$)\;
% $\mathbf{h}_{G_n} \gets$ GraphReadout($\mathbf{X}^{pool}_{n}$)\;
% $z_n, s_{n} \gets$ LSTM($\mathbf{h}_{G_n}$, $s_{n-1}$)\\
% $\hat{y}_0, \hat{y}_1 \gets$ LogSoftmax(MLP($z_n$))\;
% \uIf{$\hat{y}_1 \geq \hat{y}_0$}{ 
% \KwRet $1, s_{n}$\;}
% \uElseIf{$\hat{y}_0 > \hat{y}_1$}{
% \KwRet $0, s_{n}$\;}
% }
% \Fn{\FMain{}}{
% \While{true}{
% $I_n \gets$ Camera($n$)\;
% output, $s_n \gets$ sg2vec($I_n$, $s_{n-1}$)\;
% $n \gets n+1$\;
% \textbf{yield} output\;
% }
% }
% \caption{Collision Prediction Algorithm}
% \label{alg:collprediction}
% \end{algorithm}
% \vspace{-2em}

%% file: tex/results.tex
\section{Experimental Results}
\label{sec:exp}
This section provides extensive experimental results to demonstrate \textsc{sg2vec}'s performance, efficiency, and transferability compared to the state-of-the-art collision prediction model, DPM \cite{strickland2018deep}. 
For \textsc{sg2vec}, we used 2 MR-GCN layers, each of size 64, one \textit{SAGPooling} layer with a pooling ratio of 0.25, one \textit{add-readout} layer, one LSTM layer with hidden size 20, one MLP layer with an output of size 2, and a LogSoftmax to generate the final confidence value for each class. 
For the DPM, we followed the architecture used in \cite{strickland2018deep}, which uses one 64x64x5 Convolutional LSTM (ConvLSTM) layer, one 32x32x5 ConvLSTM layer, one 16x16x5 ConvLSTM layer, one MLP layer with output size 64, one MLP layer with output size 2, and a Softmax to generate the final confidence value.
For both models, we used a dropout of 0.1 and ReLU activation. The learning rates were 0.00005 for \textsc{sg2vec} and 0.0001 for DPM.
We ran the experiments shown in Sections \ref{subsec:coreperf} and \ref{subsec:transfer} on a Windows PC with an AMD Ryzen Threadripper 1950X processor, 16 GB RAM, and an Nvidia GeForce RTX 2080 Super GPU.

\begin{figure*}[ht]
    \centering
    \includegraphics[width=\textwidth, clip, trim=0 250 200 0]{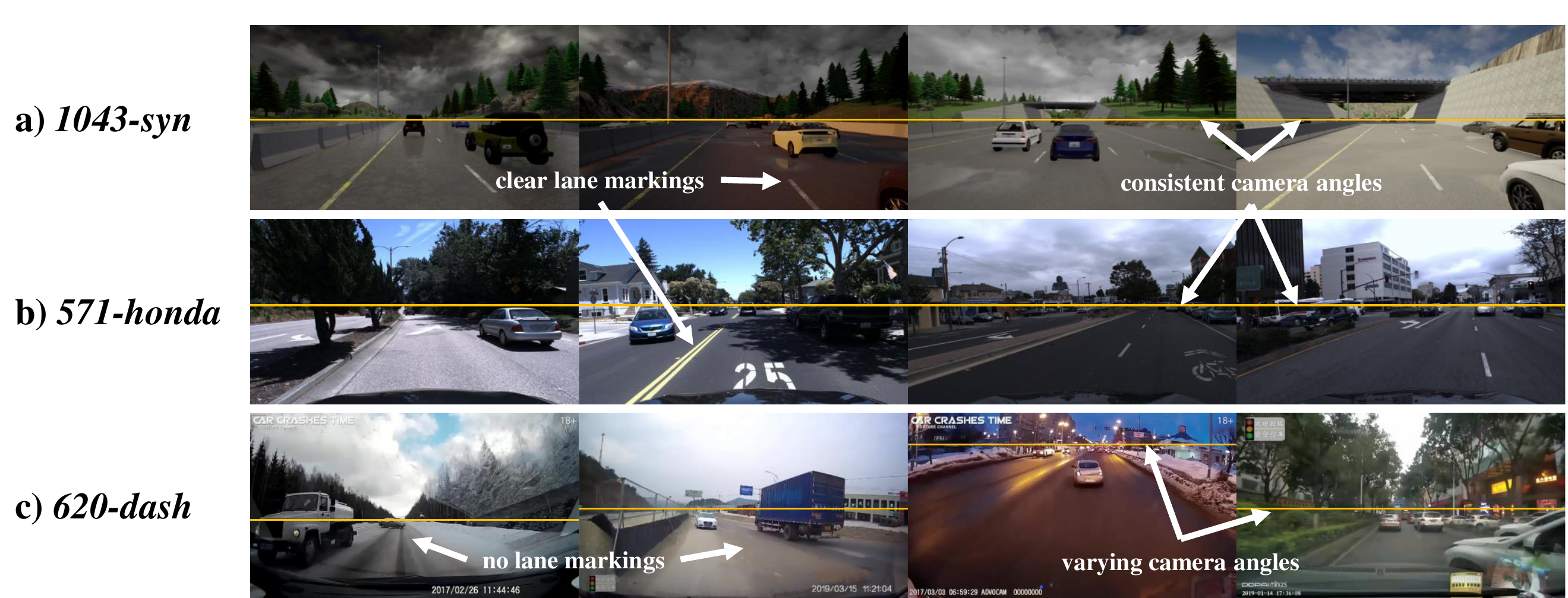}
    \caption{Examples of driving scenes from  our a) synthetic datasets, b) typical real-world dataset, and c) complex real-world dataset. In a), all driving scenes occur on highways with the same camera position and clearly defined road markings; lighting and weather are dynamically simulated in CARLA. In b) driving scenes occur on multiple types of clearly marked roads but lighting, camera angle, and weather are consistent across scenes. c) contains a much broader range of camera angles as well as more diverse weather and lighting conditions, including rain, snow, and night-time driving; it also contains a large number of clips on unpaved or unmarked roadways, as shown.}
    \label{fig:dataset}
\end{figure*}

\subsection{Dataset Preparation}
We prepared three types of datasets for our experiments: (i) synthesized datasets, (ii) a typical real-world driving dataset, and (iii) a complex real-world driving dataset. Examples from each dataset are shown in Figure \ref{fig:dataset}.
Our synthetic datasets focus on the highway lane change scenario as it is a common AV task. 
To evaluate the transferability of each model from synthetic datasets to real-world driving, we prepared a typical real-world dataset containing lane-change driving clips.
Finally, we prepared the complex real-world driving dataset to evaluate each model's performance on a challenging dataset containing a broad spectrum of collision types, road conditions, and vehicle maneuvers. 
All datasets were collected at a 1280x720 resolution, and each clip spans 1-5 seconds. 

\subsubsection{Synthetic Datasets}
To synthesize the datasets, we developed a tool using CARLA~\cite{dosovitskiy2017carla}, an open-source driving simulator, and CARLA Scenario Runner\footnote{\url{https://github.com/carla-simulator/scenario_runner}} to generate lane change video clips with/without collisions.
%CARLA is an open-source driving simulator that can simulate a wide range of road scenarios and allows users to control vehicles in either manual or autopilot mode.
%The CARLA Scenario Runner enables users to run scripted sequences of driving actions. 
% We modified the user script in CARLA to randomly select one autonomous car, switch its mode to manual mode, and then utilize the Scenario Runner's function to make the vehicle change lanes.
%Our data processing tool allows us to synthesize specific scenarios (such as lane changes) directly instead of extracting them from long driving clips. 
We generated a wide range of simulated lane changes with different numbers of cars, pedestrians, weather and lighting conditions, etc. 
We also customized each vehicle's driving behavior, such as their intended speed, probability of ignoring traffic lights, or the chance of avoiding collisions with other vehicles. 
We generated two synthetic datasets: a \textit{271-syn} dataset and a \textit{1043-syn} dataset, containing 271 and 1,043 video clips, respectively. These datasets have no-collision:collision label distributions of 6.12:1 and 7.91:1, respectively.
In addition, we sub-sampled the \textit{1043-syn} dataset to create \textit{306-syn}: a balanced dataset that has a 1:1 distribution. 
Our synthetic \textit{scene-graph} datasets\footnote{\url{https://dx.doi.org/10.21227/c0z9-1p30}} and our source code\footnote{\url{https://github.com/AICPS/sg-collision-prediction}} are open-source and available online.

\subsubsection{Typical Real-World Driving Dataset}
This dataset, denoted as \textit{571-honda}, is a subset of the Honda Driving Dataset (HDD) \cite{ramanishka2018toward} containing 571 lane-change video clips from real-world driving with a distribution of 7.21:1. The HDD was recorded on the same vehicle during mostly safe driving in the California Bay Area.

\subsubsection{Complex Real-World Driving Dataset}
Our complex real-world driving dataset, denoted as \textit{620-dash}, contains very challenging real-world collision scenarios drawn from the Detection of Traffic Anomaly dataset \cite{yao2020dota}. This dataset contains a wide range of drivers, car models, driving maneuvers, weather/road conditions, and collision types, as recorded by on-board dashboard cameras. Since the original dataset contains only collision clips, we prepared \textit{620-dash} by splitting each clip in the original dataset into two parts: (i) the beginning of the clip until 1 second before the collision, and (ii) from 1 second before the collision until the end of the collision. We then labeled part (i) as `no-collision' and part (ii) as `collision.' The \textit{620-dash} dataset contains 315 collision video clips and 342 non-collision driving clips.

\subsubsection{Labeling and Pre-Processing}
We labeled the synthetic datasets and the \textit{571-honda} dataset using human annotators. The final label assigned to a clip is the average of the labels assigned by the human annotators rounded to 0 (no collision) and 1 (collision/near collision). Each frame in a video clip is given a label identical to the entire clip's label to train the model to identify the preconditions of a future collision.

For \textsc{sg2vec}, all the datasets were pre-processed using the \textit{scene-graph} extraction pipeline mentioned in Section~\ref{sec:extraction} to construct the \textit{scene-graphs} for each video clip.
For a given sequence, \textsc{sg2vec} can leverage the full history of prior frames for each new prediction.
For the DPM, the datasets were pre-processed to match the input format used in its original implementation \cite{strickland2018deep}. 
Thus, the DPM uses 64x64 grayscale versions of the clips in the datasets turned into sets of sub-sequences $J_n$ for a clip of length $l$ defined as follows.
% \blue{
\begin{equation}
    J_n = \{I_n, I_{n+1}, I_{n+2}, I_{n+3}, I_{n+4}\}, \text{for } n \in [1, l - 4]
\end{equation}

Since DPM only uses five prior frames to make each prediction, we also present results for \textsc{sg2vec} using the same length of history, denoted as \textsc{sg2vec} (5-frames) in the results. % \red{check if n should start from 0 or start from 1. Originally we wrote 1 to align with the problem formulation better.}

\subsection{Collision Prediction Performance}
\label{subsec:coreperf}

We evaluated \textsc{sg2vec} and the DPM using classification accuracy, area under the ROC curve (AUC) \cite{bradley1997use}, and Matthews Correlation Coefficient (MCC) \cite{chicco2020advantages}. 
% MCC is considered a balanced measure of performance for binary classification even on datasets with significant class imbalances. Its performance is determined by the difference between correct and incorrect predictions for both classes divided by a normalization term that bounds the score between -1.0 and 1.0, where 1.0 corresponds to a perfect classifier, 0.0 to a random classifier, and -1.0 to an always wrong classifier. 
MCC is considered a balanced measure of performance for binary classification even on datasets with significant class imbalances. The MCC score outputs a value between -1.0 and 1.0, where 1.0 corresponds to a perfect classifier, 0.0 to a random classifier, and -1.0 to an always incorrect classifier.
Although class re-weighting helps compensate for the dataset imbalance during training, classification accuracy is typically less reliable for imbalanced datasets, so the primary metric we use to compare the models is MCC.
We used stratified 5-fold cross-validation to produce the final results shown in Table \ref{tab:results} and Figure \ref{fig:transfer}.

\begin{table}
    \centering
    \begin{tabular}{c c c c c c}
    \hline
    Dataset & Model & Accuracy & AUC & MCC\\\hline
271-syn & \textsc{sg2vec} (5-frames) & \textbf{0.8979} & \textbf{0.9541} & \textbf{0.5362}\\
271-syn & \textsc{sg2vec} & 0.8812 & 0.9457 & 0.5145 \\
271-syn & DPM & 0.8733 & 0.8939 & 0.2160  \\\hline
306-syn & \textsc{sg2vec} (5-frames) &0.7946& 0.8653& 0.5790\\
306-syn & \textsc{sg2vec} & \textbf{0.8372} & \textbf{0.9091} & \textbf{0.6812}  \\
306-syn & DPM & 0.6846  & 0.6881 & 0.3677 \\\hline
1043-syn & \textsc{sg2vec} (5-frames) & \textbf{0.9142}& \textbf{0.9623}& 0.5323\\
1043-syn & \textsc{sg2vec} & 0.9095 & 0.9477 & \textbf{0.5385}  \\
1043-syn & DPM & 0.8834  & 0.9175 & 0.2912\\\hline
620-dash & \textsc{sg2vec} (5-frames) & \green{0.6534} & \green{0.7113} & \green{0.3053}\\
620-dash & \textsc{sg2vec} & \green{\textbf{0.7007}} & \green{\textbf{0.7857}} & \green{\textbf{0.4017}} \\
620-dash & DPM & 0.4890  & 0.4717 & -0.0366  \\\hline
    \end{tabular}
    \caption{Classification accuracy, AUC, and MCC for \textsc{sg2vec} (Ours) and DPM.}
    \label{tab:results}
    %\vspace{-1.5em}
\end{table}

\subsubsection{Synthetic Datasets} 
The performance of \textsc{sg2vec} and the DPM on our synthetic datasets is shown in Table \ref{tab:results}. 
We find that our \textsc{sg2vec} achieves higher accuracy, AUC, and MCC on every dataset, even when only using five prior frames as input.
In addition to predicting collisions more accurately, \textsc{sg2vec} also infers \textbf{5.5x} faster than the DPM on average. We attribute this to the differences in model complexity between our \textsc{sg2vec} architecture and the much larger DPM model. 
Interestingly, \textsc{sg2vec} (5-frames) achieves slightly better accuracy and AUC than \textsc{sg2vec} on the imbalanced datasets and slightly lower overall performance on the balanced datasets. This is likely because the large number of safe lane changes in the imbalanced datasets adds noise during training and makes the full-history version of the model perform slightly worse. However, the full model can learn long-tail patterns for collision scenarios and performs better on the balanced datasets.

The DPM achieves relatively high accuracy and AUC on the imbalanced \textit{271-syn} and \textit{1043-syn} datasets, but suffers significantly on the balanced \textit{306-syn} dataset. This drop indicates that the DPM could not identify the minority class (collision) well and tended to over-predict the majority class (no-collision).
In terms of MCC, the DPM scores higher on the \textit{306-syn} dataset than what it scores on the other datasets. %, which is comparable to the front-camera MCC score of 0.4311 reported in \cite{strickland2018deep}.
This result is because the \textit{306-syn} dataset has a balanced class distribution compared to the other datasets, which could enable the DPM to improve its prediction accuracy on the collision class. %and lower accuracy/AUC on the \textit{306-syn} dataset compared to the other dataset because it had an equal number of examples for both classes, improving its positive-class (collision) performance at a detriment to its negative-class (no-collision) performance.

In contrast, the \textsc{sg2vec} methodology performs well on both balanced and imbalanced synthetic datasets with an average MCC of \textbf{0.5860}, an average accuracy of \textbf{87.97\%}, and an average AUC of \textbf{0.9369}.
Since MCC is scaled from -1.0 to 1.0, \textsc{sg2vec} achieves a \textbf{14.72\%} higher average MCC score than the DPM model.

The results from our \textsc{sg2vec} ablation study are shown in Table \ref{tab:ablation} and support our hypothesis that both spatial modeling with MRGCN and temporal modeling with LSTM are core to \textsc{sg2vec}'s collision prediction performance. However, the MRGCN appears to be slightly more critical to performance than the LSTM. Interestingly the choice of pooling layer (no pooling, Top-K pooling, or SAG Pooling) does not seem to significantly affect performance at this task as long as LSTM is used; when no LSTM is used SAG Pooling presents a clear performance improvement.

\subsubsection{Complex Real-World Dataset} 
The performance of both the models significantly drops on the highly complex real-world \textit{620-dash} dataset due to the variations in the driving scenes and collision scenarios. \green{This drop is to be expected as this dataset contains a wide range of driving actions, road environments, and collision scenarios, increasing the difficulty of the problem significantly. We took several steps to try and address this performance drop. First, we improved the birds-eye view (BEV) calibration on this dataset in comparison to the other datasets.
Since the varying camera angles and road conditions in this dataset impact our ability to properly calibrate \textsc{sg2vec}'s BEV projection in a single step, we created custom BEV calibrations for each clip in the dataset, which improved performance somewhat. However, as shown in Figure 4c, a significant part of the dataset consists of driving clips on roads without any discernible lane markings, such as snowy, unpaved, or unmarked roadways. These factors make it challenging to correlate known fixed distances (i.e., the width and length of lane markings) with the projections of these clips. To further improve performance on this particular dataset, we performed extensive architecture and hyperparameter tuning. 
We found that, with one MRGCN layer of size 64, one LSTM layer with hidden size 100, no SAGPooling layer, and a high learning rate and batch size, we achieved significantly better performance than the model architecture discussed in the beginning of Section \ref{sec:exp} (2 MRGCN layers of size 64, one LSTM layer with hidden size 20, and a SAGPooling layer with a keep ratio of 0.5). We believe this indicates that the temporal features of each clip in this dataset are more closely related to collision likelihood than the spatial features in each clip. As a result, the additional spatial modeling components were likely causing overfitting and skewing the spatial embedding output. The spatial embeddings remained more general with a simpler spatial model (1 MRGCN and no SAGPooling). This change, combined with using a larger LSTM layer, enabled the model to capture more temporal features when modeling each clip and better generalize to the testing set. Model performance on this dataset and similar datasets could likely be improved by acquiring more consistent data via higher-resolution cameras with fixed camera angles and more accurate BEV projection approaches. However, as collisions are rare events, there are little to no datasets containing real-world collisions that meet these requirements.   
Despite these limitations, \textsc{sg2vec} outperforms the DPM model by a significant margin, achieving \textbf{21.17\%} higher accuracy, \textbf{31.40\%} higher AUC, and a \textbf{21.92\%} higher MCC score.} Since DPM achieves a negative MCC score, its performance on this dataset is worse than that of a random classifier (MCC of 0.0). \green{Consistent with the synthetic dataset results, \textit{sg2vec} using all frames performs better on the balanced \textit{620-dash} dataset than \textsc{sg2vec} (5-frames). 
Overall, these results show that, on very challenging and complex real-world driving scenarios, \textsc{sg2vec} can perform much better than the current state-of-the-art.}

\subsubsection{Time of Prediction} 
Since collision prediction is a time-sensitive problem, we evaluated our methodology and the DPM on their average time-of-prediction (ATP) for video clips containing collisions. To calculate the ATP, we recorded the first frame index in each collision clip when the model correctly predicts that a collision would occur. We then averaged these indices and compared them with the average collision video clip length. Essentially, ATP gives an estimate of how early each model can predict a future collision. These results are shown in Table \ref{tab:atp}. On the \textit{1043-syn} dataset, \textsc{sg2vec} achieves 0.1725 for the ratio of the ATP and the average sequence length while the DPM achieves a ratio of 0.2382, indicating that \textsc{sg2vec} predicts future collisions \textbf{39.07\%} earlier than the DPM on average. In the context of real-world collision prediction, the average sequence in the \textit{1043-syn} dataset represents 1.867 seconds of data. Thus, our methodology predicted collisions \textbf{122.7} milliseconds earlier than DPM on average. This extra time can be critical for ensuring that the AV avoids an impending collision.

\begin{table}
    \centering
    \begin{tabular}{ c c c c c }
        \hline
        Dataset &  Model & ATP & Avg. Seq. Len. & Ratio\\\hline
        \textit{271-syn}   & \textbf{\textsc{sg2vec} (Ours)}  & 10.004  & 33.920 & \textbf{0.2949}\\
        \textit{271-syn}   & DPM   & 17.399 & 32.899 & 0.5289\\
        \textit{1043-syn}  & \textbf{\textsc{sg2vec} (Ours)}   & 6.442 & 37.343 & \textbf{0.1725}\\
        \textit{1043-syn}  & DPM   & 9.018 & 37.856 & 0.2382\\\hline
    \end{tabular}
    \caption{Average time of prediction (ATP) for collisions.} 
    \label{tab:atp}
    \vspace{-1.0em}
\end{table}
\begin{table}
    \centering
    \begin{tabular}{p{40pt} p{32pt} p{30pt} p{30pt} c c}
       \hline
       
       Experiment & Spatial Model & Graph Pooling & Temporal Model & Acc. & MCC \\\hline
       \multirow{4}{8pt}{Ablation Study} 
        & MLP & none & none & 0.7605 & 0.2612\\
        & MLP & none & LSTM & 0.7660 & 0.2874\\
        & MRGCN & none & none & 0.8605 & 0.4792\\
        & MRGCN & none & LSTM & \textbf{0.8931} & \textbf{0.5561}\\\hline
       \multirow{4}{8pt}{Graph Attn. and Pooling} 
        & MRGCN & Top-K & none & 0.8288 & 0.3458\\
        & MRGCN & SAGPool & none & 0.8738 & 0.5032\\
        & MRGCN & Top-K & LSTM & \textbf{0.9014} & \textbf{0.5565}\\
        & MRGCN & SAGPool & LSTM & \textbf{0.9076} & \textbf{0.5407}\\
         \hline
    \end{tabular}
    \caption{\textsc{sg2vec} ablation study on the \textit{1043-syn} dataset.}
    \label{tab:ablation}
    \vspace{-1.0em}
\end{table}

\subsection{Transferability From Synthetic to Real-World Datasets}
\label{subsec:transfer}
\begin{figure}
    \centering
    \includegraphics[trim=293 195 290 175, clip, width=\linewidth]{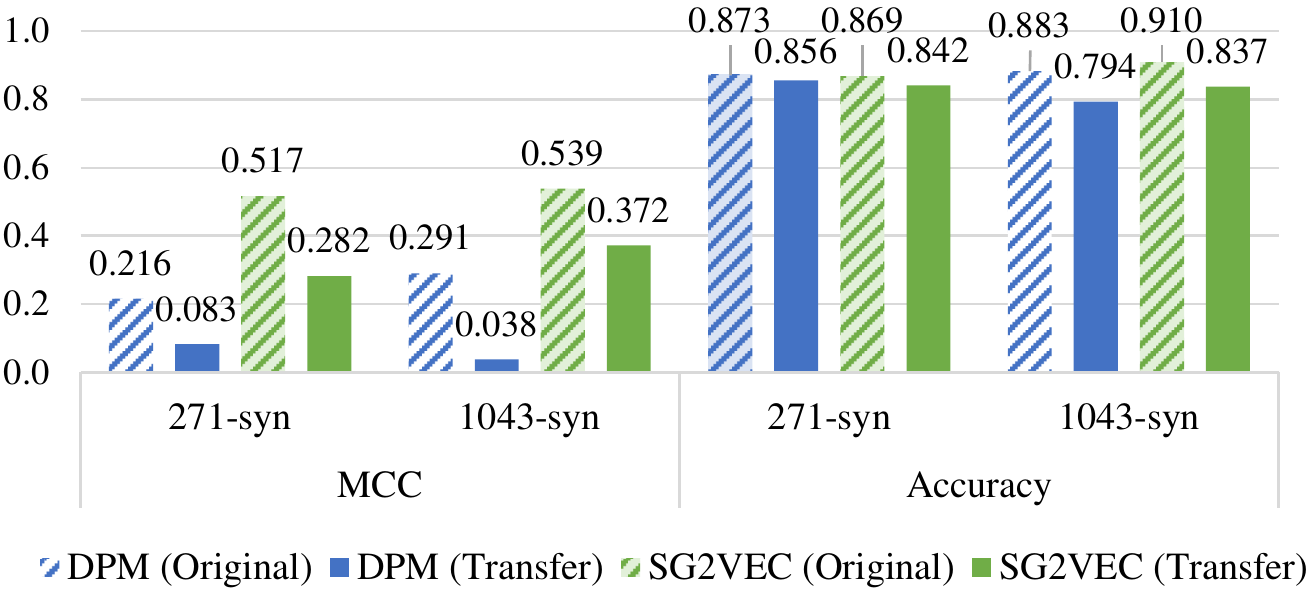}
    \caption{Performance after transferring the models trained on synthetic \textit{271-syn} and \textit{1043-syn} datasets to the real-world \textit{571-honda} dataset.}
    \label{fig:transfer}
    %\vspace{-1.5em}
\end{figure}

The collision prediction models trained on simulated datasets must be transferable to real-world driving as it can differ significantly from simulations. 
To evaluate each model's ability to transfer knowledge, we trained each model on a synthetic dataset before testing it on the \textit{571-honda} dataset. No additional domain adaptation was performed.
We did not evaluate transferability to the \textit{620-dash} dataset because it contains a wide range of highly dynamic driving maneuvers that were not present in our synthesized datasets. As such, evaluating transferability between our synthesized datasets and the 620-dash dataset would yield poor performance and would not provide insight.
Figure \ref{fig:transfer} compares the accuracy and MCC for both the models on each training dataset and the \textit{571-honda} dataset after transferring the trained model.

%We find that the MCC score for \textsc{sg2vec} drops only by 10.85\%-15.49\%  compared to the 10.94\%-19.60\% drop for the DPM model.
We observe that the \textsc{sg2vec} model achieves a significantly higher MCC score than the DPM model after the transfer, suggesting that our methodology can better transfer knowledge from a synthetic to a real-world dataset compared to the state-of-the-art DPM model. 
%Also, the \textsc{sg2vec} model's performance after transfer was found to be better than the performance of the DPM on the original synthetic datasets (see Fig. \ref{fig:transfer}).
The drop in MCC values observed for both the models when transferred to the \textit{571-honda} dataset can be attributed to the characteristic differences between the simulated and real-world datasets; the \textit{571-honda} dataset contains a more heterogeneous set of road environments, lighting conditions, driving styles, etc., so a drop in performance after the transfer is expected.
We also note that the MCC score for the \textsc{sg2vec} model trained on \textit{271-syn} dataset drops more than the model trained on the \textit{1043-syn} dataset after the transfer, likely due to the smaller training dataset size. 
Regarding accuracy, the \textsc{sg2vec} model trained on \textit{1043-syn} achieves 4.37\% higher accuracy and the model trained \textit{271-syn} dataset achieves 1.47\% lower accuracy than the DPM model trained on the same datasets. The DPM's similar accuracy after transfer likely results from the class imbalance in the \textit{571-honda} dataset.
Overall, we hypothesize that \textsc{sg2vec}'s use of an intermediate representation (i.e., \textit{scene-graphs}) inherently improves its ability to generalize and thus results in an improved ability to transfer knowledge compared to CNN-based deep learning approaches.

\subsection{Evaluation on Industry-Standard AV Hardware}
\begin{figure}
    \centering
    \includegraphics[width=0.8\linewidth]{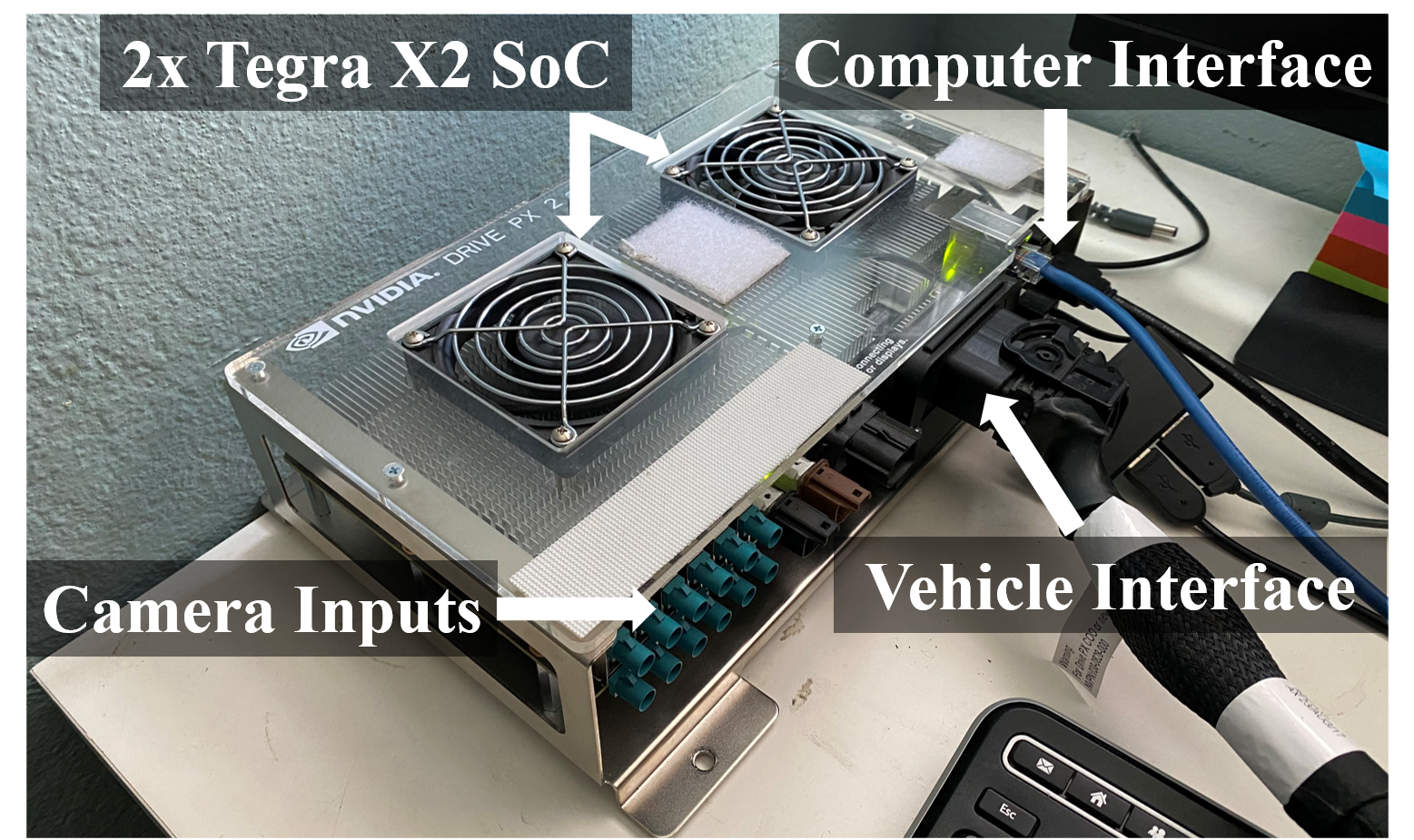}
    \caption{Our experimental setup for evaluating \textsc{sg2vec} and DPM on the industry-standard Nvidia DRIVE PX 2 hardware.}
    \label{fig:hardware}
    %\vspace{-0.5em}
\end{figure}

To demonstrate that the \textsc{sg2vec} is implementable on industry-standard AV hardware, we measured its inference time (milliseconds), model size (kilobytes), power consumption (watts), and energy consumption per frame (milli-joules) on the industry-standard Nvidia DRIVE PX 2 platform, which was used by Tesla for their Autopilot system from 2016 to 2018 \cite{px2tesla}. Our hardware setup is shown in Figure \ref{fig:hardware}.
For the inference time, we evaluated the average inference time (AIT) in milliseconds taken by each algorithm to process each frame.
We recorded power usage metrics using a power meter connected to the power supply of the PX 2. To ensure that the reported numbers only reflected each model's power consumption and not that of background processes, we subtracted the hardware's idle power consumption from the averages recorded during each test. 
For a fair comparison, we captured the metrics for the core algorithm (i.e., the \textsc{sg2vec} and DPM model), excluding the contribution from data loading and pre-processing.
Both models were run with a batch size of 1 to emulate the real-world data stream where images are processed as they are received. 
For comparison, we also show the AIT on a PC for the two models.

Our results are shown in Table \ref{tab:hw_metrics}. \textsc{sg2vec} performs inference \textbf{9.3x} faster than the DPM on the PX 2 with an \textbf{88.0\%} smaller model and \textbf{32.4\%} less power, making it undoubtedly more practical for real-world deployment. Our model also uses \textbf{92.8\%} less energy to process each frame, which can be beneficial for electric vehicles with limited battery capacity.
With an AIT of 0.4828 ms, \textsc{sg2vec} can theoretically process up to 2,071 frames/second (fps). In contrast, with an AIT of 4.535 ms, the DPM can only process up to 220 fps. In the context of real-world collision prediction, this means that \textsc{sg2vec} could easily support multiple 60 fps camera inputs from the AV while DPM would struggle to support more than three.

\begin{table}
    \centering
    \begin{tabular}{p{25pt} p{25pt} p{30pt} p{20pt} p{20pt} p{40pt}}
    \hline
        Model & PC AIT (ms) & PX2 AIT (ms) & Size (KB) & Power (W) & Energy/frame (mJ)\\\hline
        \textbf{\textsc{sg2vec}} & \textbf{0.2549} & \textbf{0.4828} & \textbf{331} & \textbf{2.99} & \textbf{1.44}\\
        DPM & 1.393 & 4.535 & 2,764 & 4.42 & 20.0\\
         \hline
    \end{tabular}
    \caption{Performance evaluation of inference on \textit{271-syn} on the Nvidia DRIVE PX 2.}
    \label{tab:hw_metrics}
    \vspace{-0.5em}
\end{table}

%% file: tex/conclusion.tex
%\vspace{-1.0em}
\section{Conclusion}
With our experiments, we demonstrated that our \textit{scene-graph} embedding methodology for collision prediction, \textsc{sg2vec}, outperforms the state-of-the-art method, DPM, in terms of average MCC (0.5055 vs. 0.2096), average inference time (0.255 ms vs. 1.39 ms), and average time of prediction (39.07\% sooner than DPM). Additionally, we demonstrated that \textsc{sg2vec} could transfer knowledge from synthetic datasets to real-world driving datasets more effectively than the DPM, achieving an average transfer MCC of 0.327 vs. 0.060. Finally, we showed that our methodology performs faster inference than the DPM (0.4828 ms vs. 4.535 ms) with a smaller model size (331 KB vs. 2,764 KB) and reduced power consumption (2.99 W vs. 4.42 W) on the industry-standard Nvidia DRIVE PX 2 autonomous driving platform. 
In the context of real-world collision prediction, these results indicate that \textsc{sg2vec} is a more practical choice for AV safety and could significantly improve consumer trust in AVs. 
Few works have explored graph-based solutions for other complex AV challenges such as localization, path planning, and control. These are open research problems that we reserve for future work.
%\vspace{-1.0em}

%% file: tex/author_bios.tex
%\vspace{-3em}

\begin{IEEEbiography}[
    {\includegraphics[width=1in,height=1.25in,clip,keepaspectratio]
{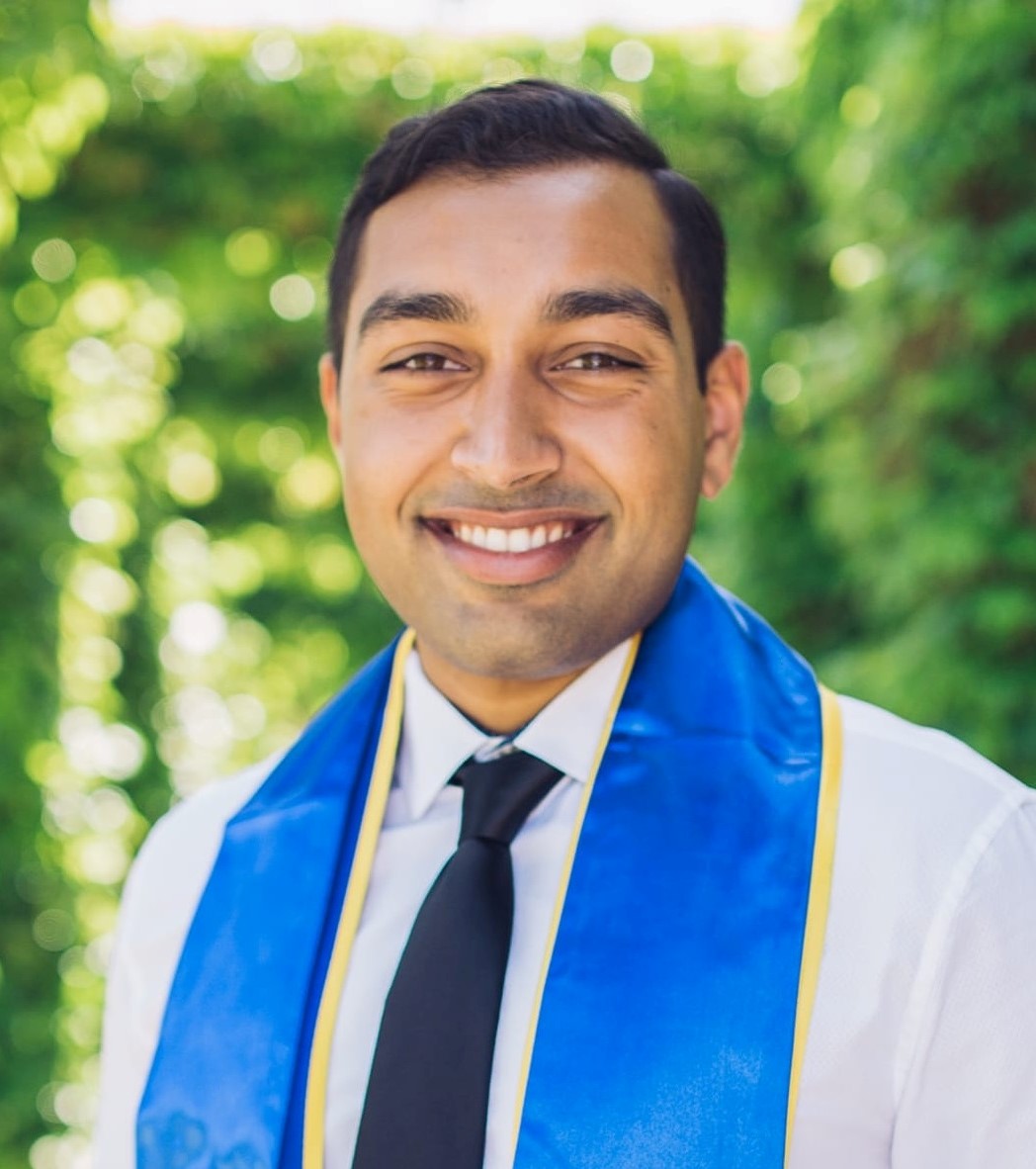}}]
{Arnav Vaibhav Malawade} received a B.S. in Computer Science and Engineering and an M.S. in Computer Engineering from the University of California Irvine (UCI) in 2018 and 2021, respectively. He is currently a Ph.D. student at UCI under the supervision of Professor Mohammad Al Faruque. His research interests include the design and security of cyber-physical systems in connected/autonomous vehicles, manufacturing, IoT, and healthcare.
\end{IEEEbiography}

%\vspace{-1em}

\begin{IEEEbiography}[
    {\includegraphics[width=1in,height=1.25in,clip,keepaspectratio]
{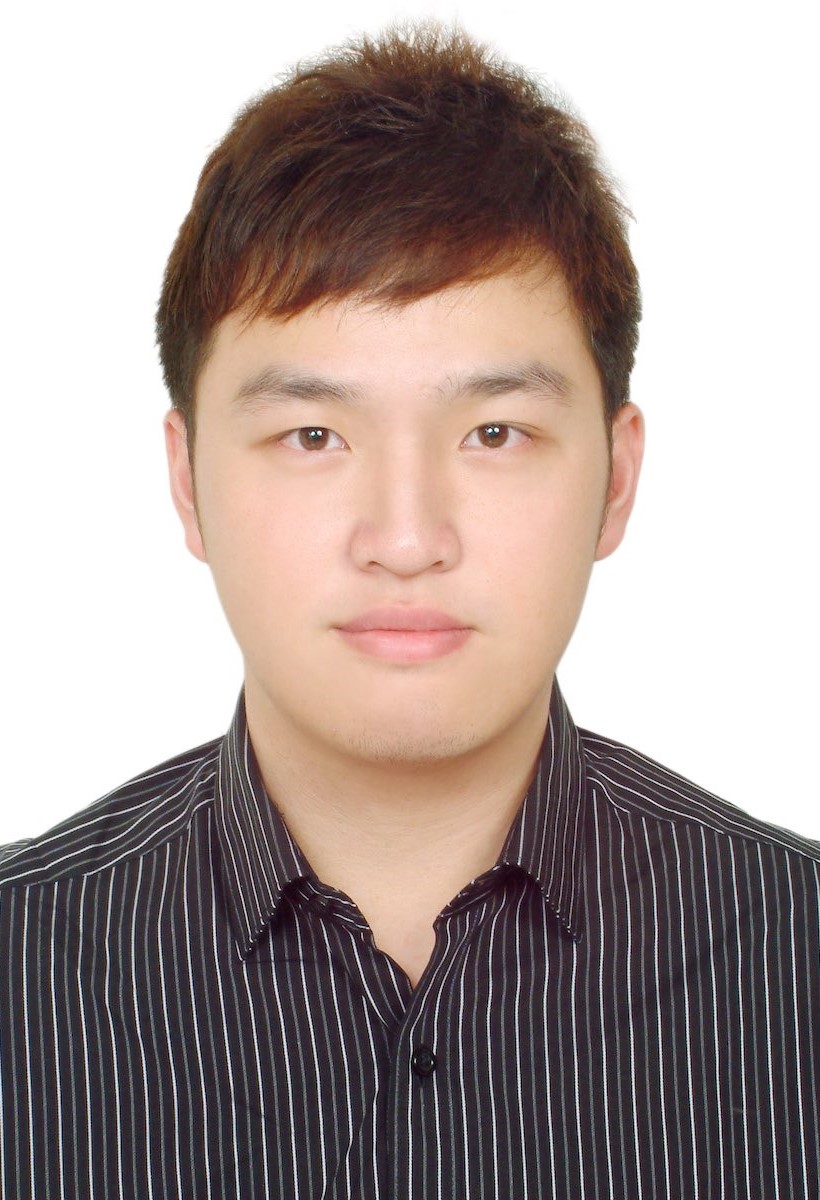}}]
{Shih-Yuan Yu} received the B.S. and M.S. degrees in Computer Science and Information Engineering from the National Taiwan University (NTU) in 2014. He worked at MediaTek for 4 years. Currently he is a Ph.D. student in the University of California, Irvine.  
Now his research interests are about design automation of embedded systems using data-driven system modeling approaches.
It covers incorporating machine learning methods to identify potential security issues in systems.
\end{IEEEbiography}

%\vspace{-1em}

\begin{IEEEbiography}
[{\includegraphics[width=1in,height=1.25in,clip,keepaspectratio]{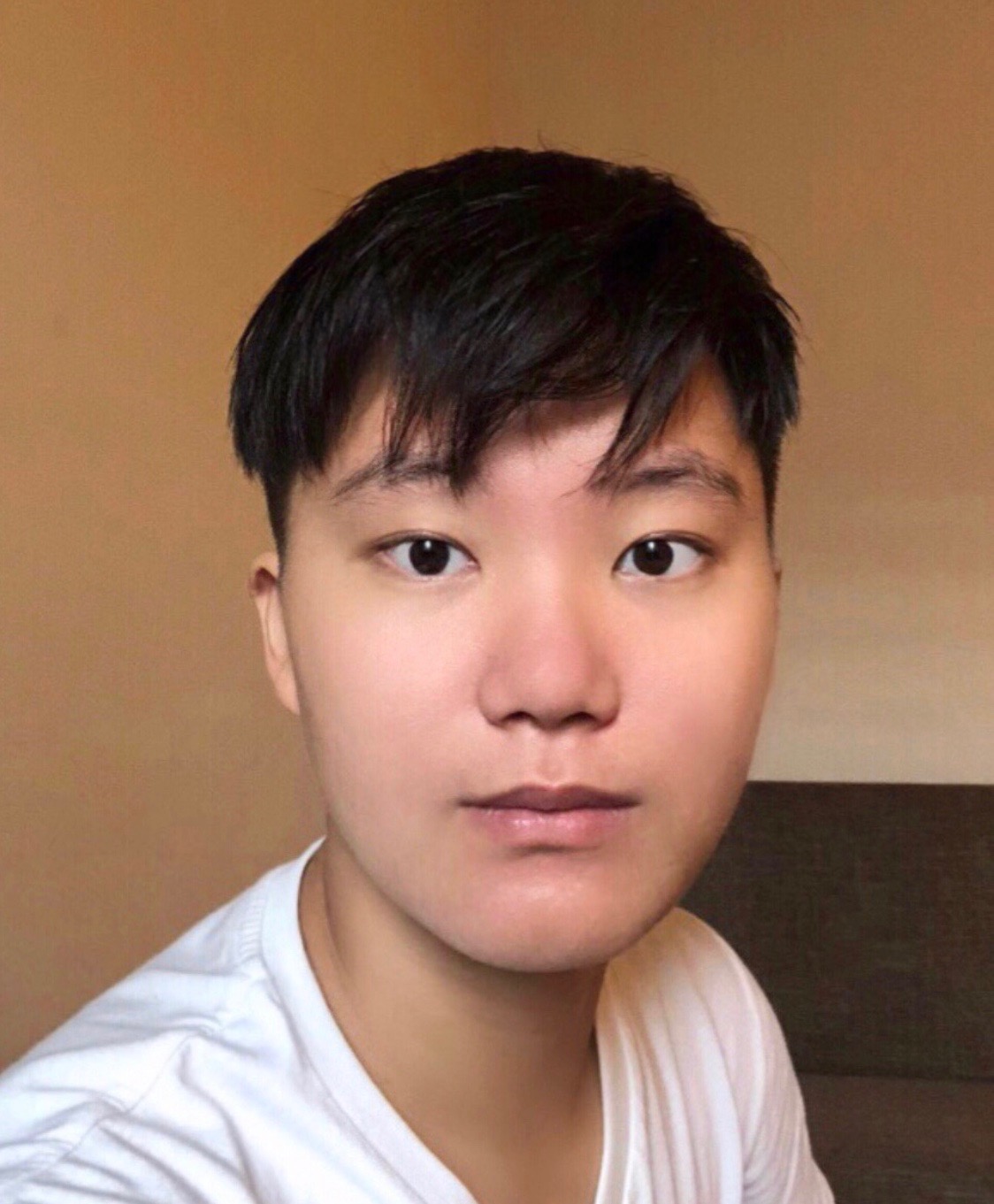}}]
{Brandon Hsu} received a B.S. degree in Computer Engineering from the University of California Irvine (UCI) in 2021. His research interests lie broadly in machine and statistical learning with current focus on perception, representation learning, and autonomous vehicles.
\end{IEEEbiography}

%\vspace{-1em}

\begin{IEEEbiography}[{\includegraphics[width=1in,height=1.25in,clip,keepaspectratio]{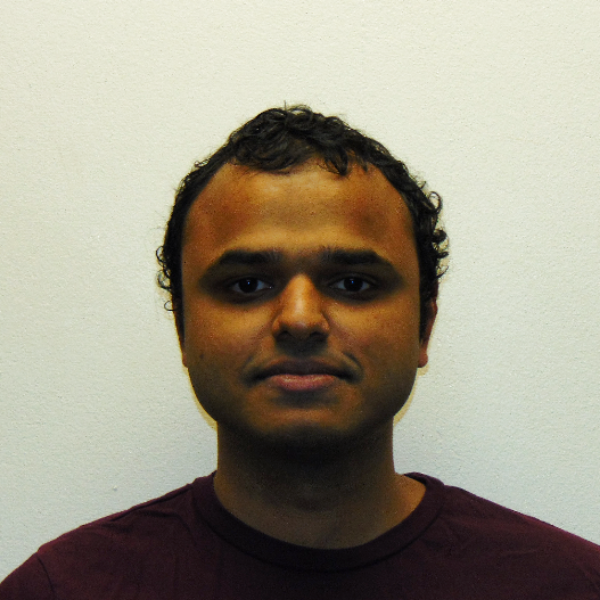}}]{Deepan Muthirayan}
is currently a Post-doctoral Researcher in the department of Electrical Engineering and Computer Science at University of California at Irvine. He obtained his Phd from the University of California at Berkeley (2016) and B.Tech/M.tech degree from the Indian Institute of Technology Madras (2010). His doctoral thesis work focused on market mechanisms for integrating demand flexibility in energy systems. Before his term at UC Irvine he was a post-doctoral associate at Cornell University where his work focused on online scheduling algorithms for managing demand flexibility. His current research interests include control theory, machine learning, topics at the intersection of learning and control, online learning, online algorithms, game theory, and their application to smart systems.
\end{IEEEbiography}

%\vspace{-1em}

\begin{IEEEbiography}
[{\includegraphics[width=1in,height=1.25in,clip,keepaspectratio]{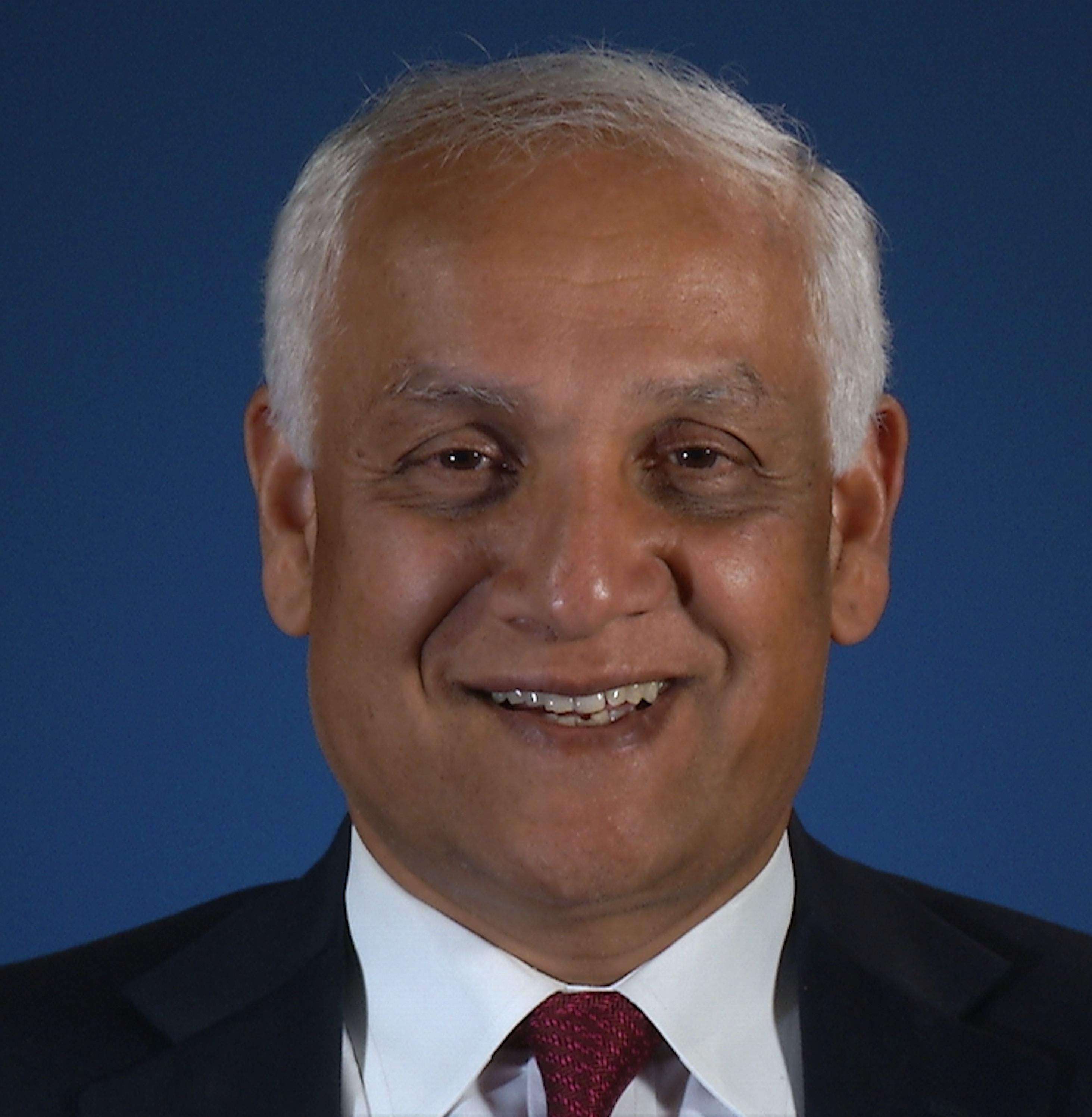}}]
{Pramod Khargonekar} received B. Tech. Degree in electrical engineering in 1977 from the Indian Institute of Technology, Bombay, India, and M.S. degree in mathematics in 1980 and Ph.D. degree in electrical engineering in 1981 from the University of Florida, respectively. He was Chairman of the Department of Electrical Engineering and Computer Science from 1997 to 2001 and also held the position of Claude E. Shannon Professor of Engineering Science at The University of Michigan.  From 2001 to 2009, he was Dean of the College of Engineering and Eckis Professor of Electrical and Computer Engineering at the University of Florida till 2016. After serving briefly as Deputy Director of Technology at ARPA-E in 2012-13, he was appointed by the National Science Foundation (NSF) to serve as Assistant Director for the Directorate of Engineering (ENG) in March 2013, a position he held till June 2016. Currently, he is Vice Chancellor for Research and Distinguished Professor of Electrical Engineering and Computer Science at the University of California, Irvine. His research and teaching interests are centered on theory and applications of systems and control. He has received numerous honors and awards including IEEE Control Systems Award, IEEE Baker Prize, IEEE CSS Axelby Award, NSF Presidential Young Investigator Award, AACC Eckman Award, and is a Fellow of IEEE, IFAC, and AAAS.
\end{IEEEbiography}

%\vspace{-1em}

\begin{IEEEbiography}
[{\includegraphics[width=1in,height=1.25in,clip,keepaspectratio]
{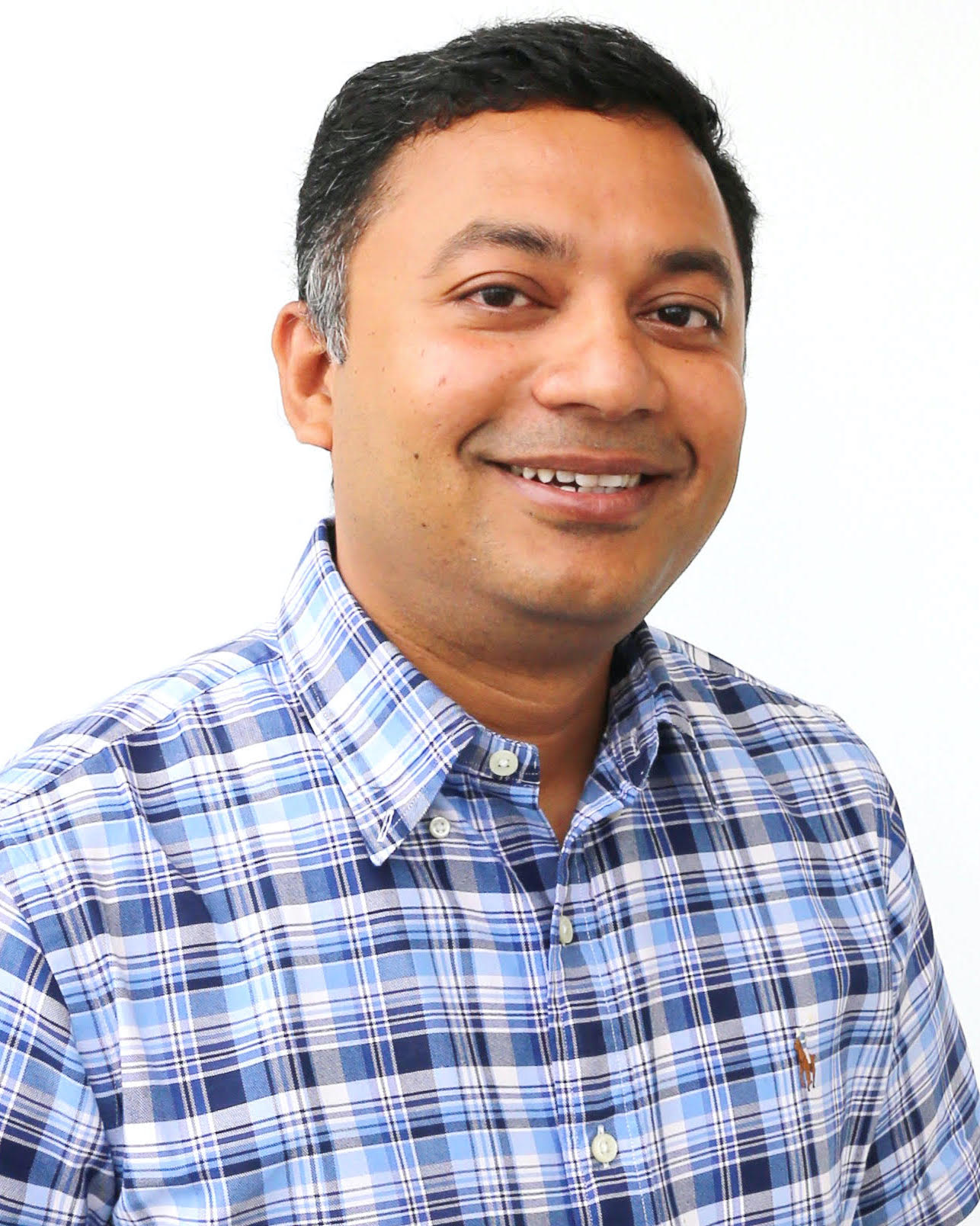}}]
{Mohammad Abdullah Al Faruque} (M’06, SM’15) received his B.Sc. degree in Computer Science and Engineering (CSE) from Bangladesh University of Engineering and Technology (BUET) in 2002, and M.Sc. and Ph.D. degrees in Computer Science from Aachen Technical University and Karlsruhe Institute of Technology, Germany in 2004 and 2009, respectively.
He is currently with the University of California Irvine (UCI) as an Associate Professor and Directing the Embedded and Cyber-Physical Systems Lab. He served as an Emulex Career Development Chair from October 2012 till July 2015. Before, he was with Siemens Corporate Research and Technology in Princeton, NJ as a Research Scientist. His current research is focused on the system-level design of embedded and Cyber-Physical-Systems (CPS) with special interest in low-power design, CPS security, data-driven CPS design, etc.
He is an ACM senior member. He is the author of 2 published books. Besides many other awards, he is the recipient of the School of Engineering Mid-Career Faculty Award for Research 2019, the IEEE Technical Committee on Cyber-Physical Systems Early-Career Award 2018, and the IEEE CEDA Ernest S. Kuh Early Career Award 2016. He is also the recipient of the UCI Academic Senate Distinguished Early-Career Faculty Award for Research 2017 and the School of Engineering Early-Career Faculty Award for Research 2017. Besides 120+ IEEE/ACM publications in the premier journals and conferences, he holds 9 US patents.
\end{IEEEbiography}

%% file: main.bbl
\begin{thebibliography}{10}
\providecommand{\url}[1]{#1}
\csname url@samestyle\endcsname
\providecommand{\newblock}{\relax}
\providecommand{\bibinfo}[2]{#2}
\providecommand{\BIBentrySTDinterwordspacing}{\spaceskip=0pt\relax}
\providecommand{\BIBentryALTinterwordstretchfactor}{4}
\providecommand{\BIBentryALTinterwordspacing}{\spaceskip=\fontdimen2\font plus
\BIBentryALTinterwordstretchfactor\fontdimen3\font minus
  \fontdimen4\font\relax}
\providecommand{\BIBforeignlanguage}[2]{{%
\expandafter\ifx\csname l@#1\endcsname\relax
\typeout{** WARNING: IEEEtran.bst: No hyphenation pattern has been}%
\typeout{** loaded for the language `#1'. Using the pattern for}%
\typeout{** the default language instead.}%
\else
\language=\csname l@#1\endcsname
\fi
#2}}
\providecommand{\BIBdecl}{\relax}
\BIBdecl

\bibitem{vom2020fail}
S.~vom Dorff, B.~B{\"o}ddeker \emph{et~al.}, ``A fail-safe architecture for
  automated driving,'' \emph{2020 Design, Automation \& Test in Europe
  Conference \& Exhibition (DATE)}, pp. 828--833, 2020.

\bibitem{bijlsma2020distributed}
T.~Bijlsma, A.~Buriachevskyi \emph{et~al.}, ``A distributed safety mechanism
  using middleware and hypervisors for autonomous vehicles,'' \emph{2020
  Design, Automation \& Test in Europe Conference \& Exhibition (DATE)}, pp.
  1175--1180, 2020.

\bibitem{NTSB2018}
{National Transportation Safety Board}, ``{Collision Between a Sport Utility
  Vehicle Operating With Partial Driving Automation and a Crash Attenuator},''
  National Transportation Safety Board, Tech. Rep. {NTSB/HAR-20/01}, 2020.

\bibitem{NTSB2019}
------, ``{Collision Between Car Operating with Partial Driving Automation and
  Truck-Tractor Semitrailer},'' National Transportation Safety Board, Tech.
  Rep. {NTSB/HAB-20/01}, 2020.

\bibitem{NTSB2019uber}
------, ``{Collision between vehicle controlled by developmental automated
  driving system and pedestrian},'' National Transportation Safety Board, Tech.
  Rep. {NTSB/HAR-19/03}, 2019.

\bibitem{deloitte2020}
{Deloitte Development LLC}, ``{Global Automotive Consumer Study - North
  America},'' Deloitte Development LLC, Tech. Rep., 2020.

\bibitem{mueller2020humanlike}
A.~S. Mueller, J.~B. Cicchino, and D.~S. Zuby, ``What humanlike errors do
  autonomous vehicles need to avoid to maximize safety?'' \emph{Journal of
  Safety Research}, 2020.

\bibitem{schoettle2015preliminary}
B.~Schoettle and M.~Sivak, ``A preliminary analysis of real-world crashes
  involving self-driving vehicles,'' University of Michigan Transportation
  Research Institute, Tech. Rep. UMTRI-2015-34, 2015.

\bibitem{xu2019statistical}
C.~Xu, Z.~Ding \emph{et~al.}, ``Statistical analysis of the patterns and
  characteristics of connected and autonomous vehicle involved crashes,''
  \emph{Journal of safety research}, vol.~71, pp. 41--47, 2019.

\bibitem{volvo2012safety}
H.~L.~D. Institute, ``Volvo collision avoidance features: initial results,''
  \emph{Highway Loss Data Institute Bulletin}, vol.~29, no.~5, 2012.

\bibitem{dahl2018collision}
J.~Dahl, G.~R. de~Campos \emph{et~al.}, ``Collision avoidance: A literature
  review on threat-assessment techniques,'' \emph{IEEE Transactions on
  Intelligent Vehicles}, vol.~4, no.~1, pp. 101--113, 2018.

\bibitem{sontges2018worst}
S.~Sontges, M.~Koschi, and M.~Althoff, ``Worst-case analysis of the
  time-to-react using reachable sets,'' in \emph{2018 IEEE Intelligent Vehicles
  Symposium (IV)}.\hskip 1em plus 0.5em minus 0.4em\relax IEEE, 2018, pp.
  1891--1897.

\bibitem{nilsson2015worst}
J.~Nilsson, A.~C. {\"O}dblom, and J.~Fredriksson, ``Worst-case analysis of
  automotive collision avoidance systems,'' \emph{IEEE Transactions on
  Vehicular Technology}, vol.~65, no.~4, pp. 1899--1911, 2015.

\bibitem{battaglia2018relational}
P.~W. Battaglia, J.~B. Hamrick \emph{et~al.}, ``Relational inductive biases,
  deep learning, and graph networks,'' \emph{arXiv preprint arXiv:1806.01261},
  2018.

\bibitem{dosovitskiy2017carla}
A.~Dosovitskiy, G.~Ros \emph{et~al.}, ``Carla: An open urban driving
  simulator,'' \emph{arXiv preprint arXiv:1711.03938}, 2017.

\bibitem{mylavarapu2020towards}
S.~Mylavarapu, M.~Sandhu \emph{et~al.}, ``Towards accurate vehicle behaviour
  classification with multi-relational graph convolutional networks,''
  \emph{arXiv preprint arXiv:2002.00786}, 2020.

\bibitem{li2020learning}
C.~Li, Y.~Meng \emph{et~al.}, ``Learning 3d-aware egocentric spatial-temporal
  interaction via graph convolutional networks,'' \emph{2020 IEEE International
  Conference on Robotics and Automation (ICRA)}, pp. 8418--8424, 2020.

\bibitem{kunze2018reading}
L.~Kunze, T.~Bruls \emph{et~al.}, ``Reading between the lanes: Road layout
  reconstruction from partially segmented scenes,'' \emph{2018 21st
  International Conference on Intelligent Transportation Systems (ITSC)}, pp.
  401--408, 2018.

\bibitem{mylavarapu2020understanding}
S.~Mylavarapu, M.~Sandhu \emph{et~al.}, ``Understanding dynamic scenes using
  graph convolution networks,'' \emph{arXiv preprint arXiv:2005.04437}, 2020.

\bibitem{yu2021scene}
S.-Y. Yu, A.~V. Malawade \emph{et~al.}, ``Scene-graph augmented data-driven
  risk assessment of autonomous vehicle decisions,'' \emph{IEEE Transactions on
  Intelligent Transportation Systems}, 2021.

\bibitem{px2tesla}
``{All new Teslas are equipped with NVIDIA's new Drive PX 2 AI platform for
  self-driving - Electrek},''
  https://electrek.co/2016/10/21/all-new-teslas-are-equipped-with-nvidias-new-drive-px-2-ai-platform-for-self-driving,
  Oct 2016, [Online; accessed 9. Nov. 2020].

\bibitem{shalev2017formal}
S.~Shalev-Shwartz, S.~Shammah, and A.~Shashua, ``On a formal model of safe and
  scalable self-driving cars,'' \emph{arXiv preprint arXiv:1708.06374}, 2017.

\bibitem{nister2019safety}
D.~Nist{\'e}r, H.-L. Lee \emph{et~al.}, ``The safety force field,''
  \emph{NVIDIA White Paper}, 2019.

\bibitem{gassmann2020integration}
B.~Gassmann, F.~Pasch \emph{et~al.}, ``Integration of formal safety models on
  system level using the example of responsibility sensitive safety and carla
  driving simulator,'' in \emph{International Conference on Computer Safety,
  Reliability, and Security}.\hskip 1em plus 0.5em minus 0.4em\relax Springer,
  2020, pp. 358--369.

\bibitem{althoff2009model}
M.~Althoff, O.~Stursberg, and M.~Buss, ``Model-based probabilistic collision
  detection in autonomous driving,'' \emph{IEEE Transactions on Intelligent
  Transportation Systems}, vol.~10, no.~2, pp. 299--310, 2009.

\bibitem{wang2019trajectory}
Y.~Wang, Z.~Liu \emph{et~al.}, ``Trajectory planning and safety assessment of
  autonomous vehicles based on motion prediction and model predictive
  control,'' \emph{IEEE Transactions on Vehicular Technology}, vol.~68, no.~9,
  pp. 8546--8556, 2019.

\bibitem{zhang2020surrounding}
L.~Zhang, W.~Xiao \emph{et~al.}, ``Surrounding vehicles motion prediction for
  risk assessment and motion planning of autonomous vehicle in highway
  scenarios,'' \emph{IEEE Access}, vol.~8, pp. 209\,356--209\,376, 2020.

\bibitem{wang2020real}
X.~Wang, J.~Liu \emph{et~al.}, ``A real-time collision prediction mechanism
  with deep learning for intelligent transportation system,'' \emph{IEEE
  transactions on vehicular technology}, vol.~69, no.~9, pp. 9497--9508, 2020.

\bibitem{strickland2018deep}
M.~Strickland, G.~Fainekos, and H.~B. Amor, ``Deep predictive models for
  collision risk assessment in autonomous driving,'' \emph{2018 IEEE
  International Conference on Robotics and Automation (ICRA)}, pp. 1--8, 2018.

\bibitem{asgari2020pisces}
B.~Asgari, R.~Hadidi \emph{et~al.}, ``Pisces: Power-aware implementation of
  slam by customizing efficient sparse algebra,'' \emph{2020 57th ACM/IEEE
  Design Automation Conference (DAC)}, pp. 1--6, 2020.

\bibitem{baidya2020vehicular}
S.~Baidya, Y.-J. Ku \emph{et~al.}, ``Vehicular and edge computing for emerging
  connected and autonomous vehicle applications,'' \emph{2020 57th ACM/IEEE
  Design Automation Conference (DAC)}, pp. 1--6, 2020.

\bibitem{huang2020opportunistic}
C.~Huang, S.~Xu \emph{et~al.}, ``Opportunistic intermittent control with safety
  guarantees for autonomous systems,'' \emph{2020 57th ACM/IEEE Design
  Automation Conference (DAC)}, pp. 1--6, 2020.

\bibitem{yang2018graph}
J.~Yang, J.~Lu \emph{et~al.}, ``Graph r-cnn for scene graph generation,''
  \emph{Proceedings of the European conference on computer vision (ECCV)}, pp.
  670--685, 2018.

\bibitem{xu2017scene}
D.~Xu, Y.~Zhu \emph{et~al.}, ``Scene graph generation by iterative message
  passing,'' \emph{Proceedings of the IEEE conference on computer vision and
  pattern recognition}, pp. 5410--5419, 2017.

\bibitem{he2017mask}
K.~He, G.~Gkioxari \emph{et~al.}, ``Mask r-cnn,'' \emph{Proceedings of the IEEE
  international conference on computer vision}, pp. 2961--2969, 2017.

\bibitem{schlichtkrull2018modeling}
M.~Schlichtkrull, T.~N. Kipf \emph{et~al.}, ``Modeling relational data with
  graph convolutional networks,'' \emph{European Semantic Web Conference}, pp.
  593--607, 2018.

\bibitem{lee2019self}
J.~Lee, I.~Lee, and J.~Kang, ``Self-attention graph pooling,'' \emph{arXiv
  preprint arXiv:1904.08082}, 2019.

\bibitem{ramanishka2018toward}
V.~Ramanishka, Y.-T. Chen \emph{et~al.}, ``Toward driving scene understanding:
  A dataset for learning driver behavior and causal reasoning,''
  \emph{Proceedings of the IEEE Conference on Computer Vision and Pattern
  Recognition}, pp. 7699--7707, 2018.

\bibitem{yao2020dota}
Y.~Yao, X.~Wang \emph{et~al.}, ``When, where, and what? a new dataset for
  anomaly detection in driving videos,'' \emph{arXiv preprint
  arXiv:2004.03044}, 2020.

\bibitem{bradley1997use}
A.~P. Bradley, ``The use of the area under the roc curve in the evaluation of
  machine learning algorithms,'' \emph{Pattern recognition}, vol.~30, no.~7,
  pp. 1145--1159, 1997.

\bibitem{chicco2020advantages}
D.~Chicco and G.~Jurman, ``The advantages of the matthews correlation
  coefficient (mcc) over f1 score and accuracy in binary classification
  evaluation,'' \emph{BMC genomics}, vol.~21, no.~1, p.~6, 2020.

\end{thebibliography}
